\documentclass[onecolumn]{IEEEtran}

\usepackage{cite}
\usepackage{caption}
\usepackage{comment}
\ifCLASSOPTIONcompsoc
\usepackage[caption=false, font=normalsize, labelfont=sf, textfont=sf]{subfig}
\else
\usepackage[caption=false, font=footnotesize]{subfig}

\ifCLASSINFOpdf
  \usepackage[pdftex]{graphicx}
\else
  \usepackage[dvips]{graphicx}
\fi
\usepackage{amsmath}
\usepackage{amssymb}
\makeatletter

\newcommand{\Rmnum}[1]{\expandafter\@slowromancap\romannumeral #1@}
\makeatother

\newtheorem{theorem}{Theorem}
\newtheorem{lemma}{Lemma}
\newtheorem{remark}{Remark}

\newtheorem{definition}{Definition}
\newtheorem{proposition}{Proposition}
\newtheorem{corollary}{Corollary}

\usepackage{float}
\usepackage{bm}
\usepackage{algorithmic}
\usepackage{array}
\usepackage[table]{xcolor}
\usepackage{booktabs}
\usepackage{algorithm}
\usepackage{algorithmic}

\ifCLASSOPTIONcompsoc
  \usepackage[caption=false,font=normalsize,labelfont=sf,textfont=sf]{subfig}
\else
  \usepackage[caption=false,font=footnotesize]{subfig}
\fi

\ifCLASSOPTIONcaptionsoff
 \usepackage[nomarkers]{endfloat}
 \let\MYoriglatexcaption\caption
 \renewcommand{\caption}[2][\relax]{\MYoriglatexcaption[#2]{#2}}
\fi

\usepackage{url}

\begin{document}

\title{Robust Graph Learning Under\\ Wasserstein Uncertainty}

\author{Xiang~Zhang,~\IEEEmembership{Student~Member,~IEEE,}
        Yinfei~Xu,~\IEEEmembership{Member,~IEEE,}
        Qinghe~Liu,
        Zhicheng~Liu,~\IEEEmembership{Student~Member,~IEEE,}
        Jian~Lu,~\IEEEmembership{Member,~IEEE,}
        and~Qiao~Wang,~\IEEEmembership{Senior~Member,~IEEE}

\thanks{The authors are with the School of Information Science and Engineering, Southeast University, Nanjing 210096, China. (e-mail: \{xiangzhang369, yinfeixu, liuqinghe, zhichengliu, lujian1980, qiaowang\}@seu.edu.cn). (\emph{Corresponding authors: Yinfei Xu; Qiao Wang}.)

Part of this work will be submitted to NeurIPS 2021.}
}


\maketitle

\begin{abstract}
Graphs are playing a crucial role in different fields since they are powerful tools to unveil intrinsic relationships among signals. In many scenarios, an accurate graph structure representing signals is not available at all and that motivates people to learn a reliable graph structure directly from observed signals. However, in real life, it is inevitable that there exists uncertainty in the observed signals due to noise measurements or limited observability, which causes a reduction in reliability of the learned graph. To this end, we propose a graph learning framework using Wasserstein distributionally robust optimization (WDRO) which handles uncertainty in data by defining an uncertainty set on distributions of the observed data. Specifically, two models are developed, one of which assumes all distributions in uncertainty set are Gaussian distributions and the other one has no prior distributional assumption. Instead of using interior point method directly, we propose two algorithms to solve the corresponding models and show that our algorithms are more time-saving. In addition, we also reformulate both two models into Semi-Definite Programming (SDP), and illustrate that they are intractable in the scenario of large-scale graph. Experiments on both synthetic and real world data are carried out to validate the proposed framework, which show that our scheme can learn a reliable graph in the context of uncertainty.

\end{abstract}


%
\IEEEpeerreviewmaketitle

\section{Introduction}
Graphs are widely employed to characterize structured data in signal processing, machine learning and statistics since intrinsic information of structured data residing on topologically complicated domain can be flexibly represented by graph\cite{thanou2017learning}. Specifically, vertices in graph represent data entities and edges represent affinity relations between these entities\cite{dong2019learning}. A variety of fields witness the applications of graph-structured data, including urban science, social networks and meteorology, etc. Among these applications, many models first define graphs using prior knowledge which, however, is often unavailable. Furthermore, prior graphs might not accurately capture the intrinsic relationships among  vertices. Therefore, it is essential to learn an underlying graph topology directly from data on hand and then the learned graph can be used in numerous pipeline tasks.

Historically, graph learning has been a research field of concern for a long time. Abundant of researches, such as \cite{friedman2008sparse, yuan2007model, ravikumar2008model}, are committed to learning a graph from  statistical view. On the other hand, graph signal processing (GSP), an emerging research field \cite{shuman2013emerging, ortega2018graph}, is employed to handle graph learning problems from perspective of signal processing. The main characteristic of GSP based models is learning a graph with some tools generalizing classical signal processing concepts, such as sampling theorem, time-frequency analysis and filtering on graphs \cite{dong2019learning}. One of the most notable GSP based model is built on stationarity assumption\cite{mateos2019connecting, pasdeloup2017characterization, segarra2017network, egilmez2018graph}. These models hold the opinion that the observed signals are stationary over the graph to be learned and the eigenvalues of graph operators, such as Laplacian matrices, can be estimated by using sample covariance matrices of the observed data\cite{dong2019learning}. In parallel with stationarity based models, smoothness assumption also occupies an important position in GSP based graph learning models. The smoothness of graph signals means that signal values of two connected vertices in the corresponding graph with large weights tend to be similar\cite{dong2016learning}. Smoothness based models endeavor to learn a graph over which observation graph signals are the smoothest \cite{kalofolias2016learn, chepuri2017learning, huang2018rating, dong2016learning, berger2020efficient}. Here we should mention that smoothness assumption is the one we take in this paper.

\begin{figure}[t]
    \centering
	  \subfloat[]{
       \includegraphics[width=0.35\linewidth]{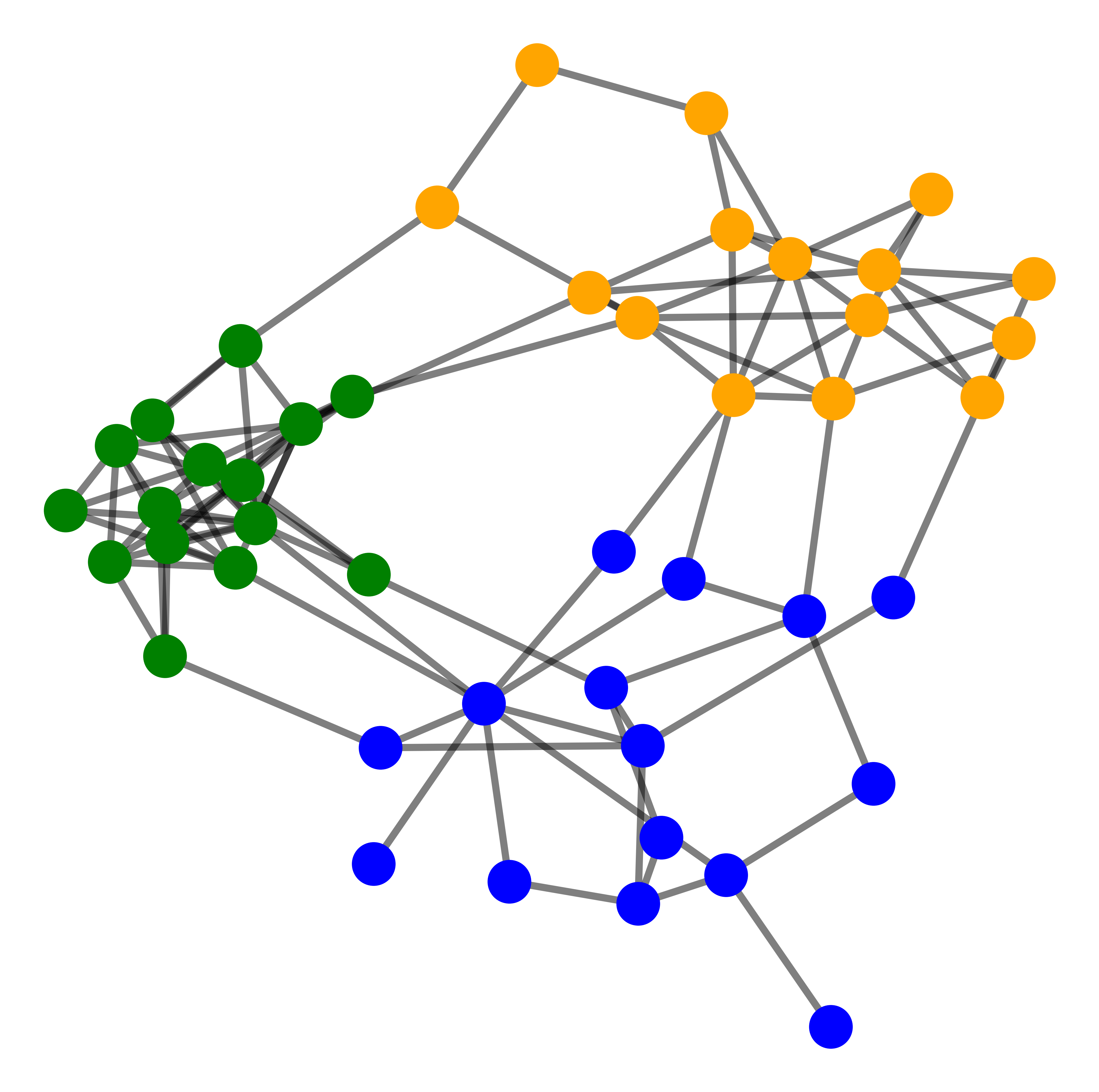}}
        \label{fig-SBM-a}
	  \quad
	  \subfloat[]{
        \includegraphics[width=0.35\linewidth]{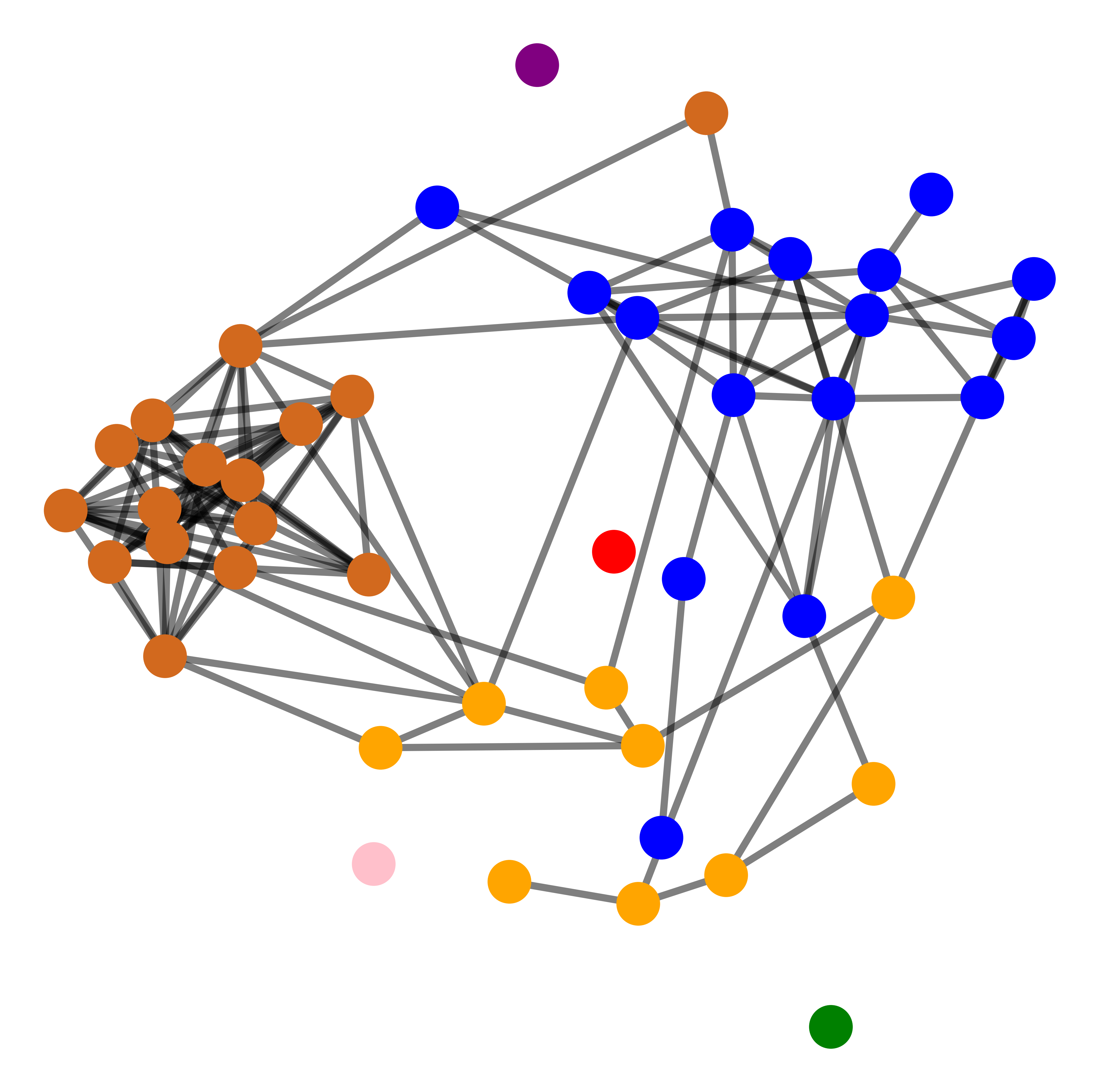}}
    \label{fig-SBM-b}\\
    	\caption{Community detection results using the learned graph. (a) The ground truth graph. (b) The learned graph under uncertainty}
    	\label{fig-SBM}
\end{figure}

While a wealth of researches are carried out on graph learning, limited to our understanding, few of them focuses on how to learn a graph when there exists uncertainty in signals. Uncertainty arises from, for example, noise measurements or limited observability of data and may cause the learned graph deviate from ground-truth, which brings troubles to pipeline tasks\cite{rahimian2019distributionally}. To better illustrate the impact of uncertainty, we might learn a graph under uncertainty using method in \cite{dong2016learning} and apply Louvain algorithm \cite{blondel2008fast} to detect clusters in the learned graph. To increase uncertainty, we only generate 50 signals for a 45 vertices graph and add noise to the generated signals. As displayed in Fig.\ref{fig-SBM}, for the learned graph, 6 clusters are detected while the groundtruth graph only contains 3 clusters. This vividly illustrates the impact of uncertainty on graph learning tasks and this motivates us to surmount uncertainty in signals.

From perspective of optimization, if we regard signals with uncertainty as unknown parameters, graph learning under uncertainty amounts to an optimization problems with unknown parameters. Typically, robust optimization \cite{beyer2007robust, ben2009robust} and stochastic optimization\cite{fouskakis2002stochastic, nemirovski2009robust} are two common tools to cope with optimization problems with unknown parameters. While robust optimization limits unknown parameters in an uncertainty set, stochastic optimization assumes that unknown parameters subject to a certain distribution. Recently, distributionally robust optimization (DRO) has been a popular tool to handle optimization problems with unknown parameters and can be viewed as a unifying framework for robust optimization and stochastic optimization \cite{rahimian2019distributionally}. Instead of directly constructing uncertainty sets with respect to the value of unknown parameters, DRO assumes that uncertainty is built on the distributions of unknown parameters \cite{rahimian2019distributionally, goh2010distributionally}. To be specific, in the context of graph learning, we have no knowledge of the true distribution of graph signals because of noise corruption or limited number of data. To this end, we construct an uncertainty set containing the true distribution with high possibility. With this uncertainty set, we expect to find the graph that minimize the risk of the worst case of all possible distribution in uncertainty set. Since DRO takes the worst case into consideration, results of DRO are conservative and robust. It is worth noting that uncertainty sets in DRO is crucial and defined on distributions, hence some distance metrics between probabilities need to be employed to construct uncertainty sets. In addition to  $\Phi$ divergence\cite{ben2013robust, duchi2016variance} and maximum mean discrepancy\cite{staib2019distributionally} based metrics, Wasserstein distance is another emerging one to measure the discrepancy between probabilities \cite{esfahani2018data}. Recall that Wasserstein distance is originated from optimal transport theory\cite{villani2008optimal} and takes into account the geometry of the space on which the distributions are defined\cite{simou2020node2coords}. In fact, Wasserstein distance based distributionally robust optimization, which is abbreviated as WDRO\cite{ gao2016distributionally, blanchet2018optimal, kuhn2019Wasserstein}, has gained success in numerous fields\cite{staib2017distributionally, shafieezadeh2015distributionally, gao2018robust}. However, to the best of our knowledge, this paper is the first work that handles uncertainty from distributional perspective exploiting WDRO in graph learning.

Our contributions in this paper may be summarized as follows:
\begin{enumerate}
\item[$\bullet$]
We first propose a graph learning framework under uncertainty using WDRO by defining uncertainty set from distributionally perspective. Specifically, two models are developed, one of which assumes all distributions in uncertainty sets are of Gaussian and the another one has no prior assumption about distributions in uncertainty set. For the second model, we prove that robustness is obtained by a regularizer whose weight equals to the radius of uncertainty set, which can be interpreted as the level of robustness. This provides interpretability for robustness under our framework.

\item[$\bullet$]
To exploit the algorithms for solving the above models, we reformulate the above models into Semi-Definite Programming (SDP) problems. Then we demonstrate that it is actually impractical to solve such SDP problems in reality, which suggests us to give up the SDP reformulations. To this end, we develop two novel algorithms respectively for Gaussian and general scenario. Instead of using convex optimization package, such as CVX exploiting interior methods, we introduce a linear operator to get rid of constraints of graph structure and use gradient descendent method to find the optimal result. The proposed algorithms are illustrated to be more time-saving.
\item[$\bullet$]
Experiments with synthetic data and real data are carried out to validate the proposed framework. These results confirm that the proposed framework is effective at learning a reliable graph in the context of uncertainty.
\end{enumerate}

\emph{Organization:} The rest of the paper is organized as follows. Section \ref{sec:problem-statement} states the problem which we focus on and gives the basic formulation of our framework. Section \ref{sec:formulation-gaussian} and section \ref{sec:formulation-general} are the models and algorithms of our framework in Gaussian and general scenario respectively. In section \ref{sec:formulation-sdp}, we reformulate these two models of section \ref{sec:formulation-gaussian} and \ref{sec:formulation-general} into SDP problems and show the infeasibility of solving procedures. Synthetic and real data experiments are carried out to validate our models in section \ref{sec:Experiments}. Finally, some concluding remarks are presented in Section \ref{sec:Conclusion}.

\emph{Notations:} Some important notations used throughout this paper are listed in Table \ref{table:notations}. Other minutiae notations will be given in detail in the corresponding sections.

	\begin{table}[t]
		\centering
		\caption{List of symbols and their meaning}
		\begin{tabular}{cc}
			\toprule
			\textbf{Symbols}  & \textbf{Meaning}  \\
			\midrule
			$\mathbf{L}$, $\mathcal{L}$, $l_{ij}$  & graph Laplacian matrix, set of graph Laplacian matrices, element of $\mathbf{L}$ \\
			$\mathbf{W}$, $\mathbf{D}$, $\mathbf{I}$  & adjancy matrix, degree matrix, identity matrix \\
			$x$, $\mathbf{x}$, $\mathbf{X}$  & signal element, signal vector, signal matrix,  \\
			$d$, $N$ & number of vertices, number of signals\\
			$\text{Tr}(\cdot)$ , $\textbf{vec}(\cdot)$ &trace operator, vectorization operator\\
			$\mathbb{E}(\cdot)$ , $\mathbf{Cov}(\cdot)$ &expectation operator, covariance operator\\
			$W(\cdot)$, $\alpha$& Wasserstein distance, Wasserstein distance type\\
			$p$, $\epsilon$ &norm type, uncertainty set radius\\
			$\gamma$, $\mathbf{\Theta}$ & dual variable, equivalent variable of $\mathbf{\mathbf{x}\mathbf{x}^{\text{T}}}$\\
			$\mathbb{S}^d$ & $d \times d$ symmetric matrix\\
			$\mathbb{S}^d_+$ & $d \times d$ positive semidefinite matrix\\
			$\mathbb{R}^d$  & $d$-dimensional real vector\\
			$\mathbb{R}^d_+$ & $d$-dimensional non-negative real vector\\
			$\mathbb{P}$, $\mathbb{Q}$ & distribution of variable $\mathbf{\mathbf{x}}$, distribution of variable $\mathbf{\Theta}$\\
			$\mathcal{P}$,$\mathcal{Q}$ & set of distribution $\mathbb{P}$, set of distribution $\mathbb{Q}$\\
			$\bm{\mu}$, $\mathbf{\Sigma}$ & mean vector of $\mathbf{x}$, covariance matrix of $\mathbf{x}$\\
			$\bm{\mu}_n$, $\mathbf{\Sigma}_n$ & empirical mean vector, empirical covariance matrix \\
			$\mathbf{1}$, $\mathbf{0}$ & vector with  entries of 1, vector with entries of 0 \\
			
			\bottomrule
		\end{tabular}
		\label{table:notations}
	\end{table}

\section{Problem Formulation}
\label{sec:problem-statement}
In this section, the smoothness based graph learning framework is firstly revisited. When there exists uncertainty in observed graph signals, the distributionally robust graph learning problem is then proposed. To measure the discrepancy of distributions in uncertainty sets, we select Wasserstein distance in this paper. Therefore,  we introduce the definition on Wasserstein distance in the last part of this section.


\subsection{Smoothness Based Graph Learning}
In this paper, we only focus on undirected graphs with nonnegative weights and no self-loops. Formally, define a graph $\mathcal{G} = \{\mathcal{V},\mathcal{E},f\}$ with $d$ vertices, where $\mathcal{V} = \{v_1,...,v_d\}$ denotes vertex set and $\mathcal{E}$ is edge set. Mapping $f$: $ \mathcal{E} \to \mathbb{R}$ allocates every edge $e$ in $\mathcal{E}$ a nonnegative real value as weight. Give a graph $\mathcal{G}$, its adjacency matrix $\mathbf{W}$ is an $d\times d$ symmetric matrix and we denote it as $\mathbf{W}\in\mathbb{S}^d$. Clearly, ${w}_{ij} ={w}_{ji} = f((v_i,v_j))$ for $i\neq j$ and ${w}_{ij} = 0 $ if $i=j$, where $(v_i,v_j)$ denotes the edge whose corresponding vertices are $v_i$ and $v_j$. Furthermore, the degree matrix $\mathbf{D}$ of $\mathcal{G}$ is defined as a $d\times d$ diagonal matrix with ${d}_{ii} = \sum_{j=1}^d \mathbf{W}_{ij}$. Based on the above definitions, Laplacian matrix can then be defined as
	\begin{equation}
    \mathbf{L} = \mathbf{D} - \mathbf{W}.
    \label{LaplacianDefinition}
    \end{equation}
Note that $\mathbf{L}$ is a positive semi-definite matrix and the set of all Laplacian matrices can be written as
	\begin{equation}
    \mathcal{L} \triangleq \{\mathbf{L}: \mathbf{L} \in \mathbb{S}_{+}^{d},\,\, \mathbf{L}\mathbf{1} = \mathbf{0},\, {l}_{ij}\leq 0 \,\, \text{for}\,\, i \neq j \}.
    \label{LaplacianSets}
    \end{equation}
In section \ref{sec:formulation-gaussian} and \ref{sec:formulation-general}, a normalized constraint $\text{Tr}(\mathbf{L}) = d$ is added to avoid trivial solution. Under this circumstance, the set of all feasible Laplacian matrices is defined as
\begin{equation}
\mathcal{L}_c \triangleq \{\mathbf{L}: \mathbf{L} \in \mathbb{S}_{+}^{d},\, \mathbf{L}\mathbf{1} = \mathbf{0}\ ,\, \text{Tr}(\mathbf{L}) = d,\, {l}_{ij}\leq 0 \,\, \text{for}\,\, i \neq j \}.
\end{equation}

A signal $\mathbf{x} = [x_1,x_2,...x_d]^\text{T}$ defined on graphs, which is named as graph signals, means that every dimension of this signals represents a vertex in the graph. The existence of edges between two vertices can be understood as that the corresponding two dimensions of the signals are related, and the weights quantify the relationships. Given $N$ observations $\mathbf{X} = [\mathbf{x}_1,...\mathbf{x}_N]$ generated from graph $\mathcal{G}$, graph learning tasks are designed to infer the topology of $\mathcal{G}$ with $d\times N $ observation matrix $\mathbf{X}$ using some prior assumptions. Recall that smoothness is the metric we adopt in this paper, it is crucial to appropriately define the smoothness of signals over graphs. Several definitions of smoothness are proposed, such as \cite{kalofolias2016learn,berger2020efficient,zhou2004regularization}, and in this paper we employ the following form.
\begin{definition}
(Smoothness). Given an observed signal $\mathbf{x}$ and the Laplacian matrix $\mathbf{L}$ of graph $\mathcal{G}$, the smoothness of $\mathbf{x}$ over $\mathcal{G}$ is defined as
\begin{equation}
        \mathbf{x }^{\text{T}}\mathbf{L}\mathbf{x } = \frac{1}{2}\sum_{i,j}{w}_{ij}(x_i-x_j)^2
        \label{smoothness}
    \end{equation}
\label{definition-smoothness}
\end{definition}
In fact, the smaller value of (\ref{smoothness}) is, the smoother the signal over the graph. And (\ref{smoothness}) explicitly indicates that if an edge weight $\mathbf{w}_{ij}$ is large, the values of the corresponding vertices of this edge are supposed to be close under smoothness assumption. This can be explained as that, under smoothness assumption, edge weights of the learned graph represent the similarity of the connected vertices. Based on the above definition, graph learning based on smoothness assumption can be formulated as
\begin{align}
    &\underset{\mathbf{L}\in \mathcal{L}}{\text{inf}}\, \sum_{i=1}^{N} \mathbf{x}^{\text{T}}_i\mathbf{L}\mathbf{x}^{\text{T}}_i +\eta\parallel\mathbf{L} \parallel_{{F}}^2 \\
    =&
    \underset{\mathbf{L}\in \mathcal{L}}{\text{inf}}\, \text{Tr}(\mathbf{X}^{\text{T}}\mathbf{L}\mathbf{X}) +\eta\parallel\mathbf{L} \parallel_{{F}}^2,
    \label{classic formulation}
\end{align}
where $\mathbf{X}$ is the $d\times N$ observation matrix. The Frobenius norm term $\parallel\mathbf{L} \parallel_{{F}}^2$  with constant $\eta$ is used to control the edge weights of the learned graph \cite{dong2016learning}. Note that
the Laplacian $\mathbf{L}$ is learned from \eqref{classic formulation} and it can represent the topology of a graph because of one-to-one relationship. We emphasize that (\ref{smoothness}) has some variant forms displayed in \cite{kalofolias2016learn}.

\subsection{Distributionally Robust Graph Learning Formulation}
We then turn our attention to how to learn a graph under smoothness assumption when there exists uncertainty in observed signals. Given $N$ $d-$dimensional observed signals $\mathbf{x}_1,\mathbf{x}_2,...\mathbf{x}_N$ with uncertainty caused by limited observability or noise corruption, we are purposed to learn a robust graph taking uncertainty into consideration. As indicated in (\ref{classic formulation}), classic formulation ignores the uncertainty in signals. If the observed signals deviate from the true ones significantly, unreliability will be brought to the learned graph.

Since the observed signals $\mathbf{x}_1,\mathbf{x}_2,...\mathbf{x}_N$ are not reliable due to uncertainty, we can take signals generated from graphs as unknown parameters. What we need to do is learning a robust graph with these unknown parameters. One natural idea is that if the distribution of $\mathbf{x}$ is available, we can minimize the expectation of (\ref{smoothness}) to eliminate the impact of unknown parameters. To be specific,  we can learn a graph by the following formulation,
\begin{equation}
\underset{\mathbf{L}\in \mathcal{L}}{\text{inf}}\,  \mathbb{E}_{\mathbb{P}_{\text{real}}} \left[ \mathbf{x}^{\text{T}}\mathbf{L}\mathbf{x} +\eta\parallel\mathbf{L} \parallel_{F}^2\right],
\label{stochatic optimization formulation}
\end{equation}
where $\mathbb{P}_{\text{real}}$ is the real distribution of $\mathbf{x}$. This formulation is called stochastic optimization (SO)\cite{rahimian2019distributionally}. In SO, we learn a Laplacian matrix with unknown parameters from the perspective of probability instead of only from the observed signals.

Unfortunately, in most situations of practical interest, $\mathbb{P}_{\text{real}}$ is hard to obtain precisely. Hence, empirical distributions, $$\mathbb{P}_{n} = {1}/{N}\sum_{i=1}^N \delta(\mathbf{x}_i),$$ are used as an alternative in many applications, where $\delta(\mathbf{x}_i)$ denotes the Dirac point mass at the $i^{\text{th}}$ training sample $\mathbf{x}_i$. Using empirical distribution, \eqref{stochatic optimization formulation} is converted into
\begin{align}
& \underset{\mathbf{L}\in \mathcal{L}}{\text{inf}}\,  \mathbb{E}_{\mathbb{P}_{n}} \left[\mathbf{x}^{\text{T}}\mathbf{L}\mathbf{x}+\eta\parallel\mathbf{L} \parallel_{{F}}^2\right] \\
= &  \underset{\mathbf{L}\in \mathcal{L}}{\text{inf}}\, \frac{1}{N}\sum_{i=1}^{N} \mathbf{x}_i^{\text{T}}\mathbf{L}\mathbf{x}_i+\eta\parallel\mathbf{L} \parallel_{{F}}^2,
\label{SAA formulation}
\end{align}
and this is called sample average approximation (SAA) \cite{esfahani2018data}. The formulation is similar to the model named SigRep of \cite{dong2016learning} but SAA is derived from probability perspective. However, when the number of observations is limited, empirical distributions may be far from true distributions, which also bring uncertainty to sample distributions. As a result, SAA tends to display a poor out-of-sample performance\cite{esfahani2018data}. To address this issue, inspired by the idea of robust optimization, we define an uncertainty set on sample distributions
\begin{equation}
\mathcal{P} = \left\{\mathbb{P}|\,\,\text{dist}(\mathbb{P},\mathbb{P}_{n}) \leq \epsilon\right\},
\label{definition-uncertainty-set}
\end{equation}
where $\text{dist}(\cdot)$ measures the distance between two distributions. The uncertainty set contains all distributions whose distance from empirical distribution $\mathbb{P}_{n}$ is less than $\epsilon$. For all distributions in uncertainty set, we define the worst case risk as
\begin{equation}
R(\mathbf{L}) = \underset{\mathbb{P}\in \mathcal{P}}{\text{sup}}\, \mathbb{E}_\mathbb{P} \left[\mathbf{x}^{\text{T}}\mathbf{L}\mathbf{x}+\eta\parallel\mathbf{L} \parallel_{{F}}^2\right].
\label{wrost case formulation}
\end{equation}
Under the viewpoint of robust optimization, we may learn a robust graph $\mathbf{L}$ by minimizing the worst case, \emph{i.e.},
\begin{equation}
\underset{\mathbf{L}\in \mathcal{L}}{\text{inf}} R(\mathbf{L})  = \underset{\mathbf{L}\in \mathcal{L}}{\text{inf}}\, \underset{\mathbb{P}\in \mathcal{P}}{\text{sup}}\, \mathbb{E}_\mathbb{P} \left[\mathbf{x}^{\text{T}}\mathbf{L}\mathbf{x}+\eta\parallel\mathbf{L} \parallel_{{F}}^2\right].
\label{basic infsup formulation}
\end{equation}

The philosophy under minimizing the worst case is that we can push down the risk under all distributions in uncertainty set, which of course includes the true distribution $\mathbb{P}_{\text{real}}$ \cite{kuhn2019Wasserstein}. For the learned $\mathbf{L}^*$, the risks of all distributions in uncertainty set are smaller than $R(\mathbf{L}^*)$. Hence, $\mathbf{L}^*$ is not sensible to the change of $\mathbb{P}$ leading to a robust result. It should be pointed out that, different from robust optimization whose uncertainty set is defined on unknown parameters directly, the uncertainty set of our formulation is defined on the unknown distributions and this is the origin of the name distributionally robust optimization (DRO). Furthermore, as shown in (\ref{definition-uncertainty-set}), $\epsilon$ determines the size of uncertainty sets. Clearly, larger $\epsilon$ implies more nuisance distributions in uncertainty set, causing $\mathbf{L}^*$ to be less sensitive to changes of $\mathbb{P}$. From this point of view, larger $\epsilon$ means the more robustness consideration. However, large $\epsilon$ may lead to  over conservative $\mathbf{L}^*$ since uncertainty sets can contain more "bad" distributions which may not be encountered at all in real life. Therefore, $\epsilon$ actually controls the level of robustness and an appropriate $\epsilon$ needs to be determined in advance.

\subsection{Wasserstein Distance}

In the definition of uncertainty set (\ref{definition-uncertainty-set}), many tools can be selected as $\text{dist}(\cdot)$, such as KL divergence, total variation metric, etc. In this paper, we select Wasserstein distance to measure the discrepancy between two distributions due to its great properties \cite{rahimian2019distributionally}. Before going into efficient algorithms to solve the Wasserstein robust optimization problem in details, we first introduce the necessary definition on type-$\alpha$ Wassertein distance formally.
\begin{definition}
(\cite{gao2016distributionally}, Definition 2 ). For any $\alpha \in [1, \infty) $, type-$\alpha$ Wasserstein distance between two probability distributions $\mathbb{P}_1$ and $\mathbb{P}_2$ on $\mathbb{R}^d$ is defined as
	\begin{equation}
	    W_{\alpha}(\mathbb{P}_1,\mathbb{P}_2) = \left( \underset{\pi \in \Pi(\mathbb{P}_1,\mathbb{P}_2)}{\text{inf}}\, \int_{\mathbb{R}^d \times \mathbb{R}^d} C(\mathbf{z}_1,\mathbf{z}_2)^{\alpha} \pi(\text{d}\mathbf{z}_1,\text{d}\mathbf{z}_2)\right)^{\frac{1}{\alpha}},
        \label{Wasserstein distance}
    \end{equation}
where $C(\cdot)$ denotes a cost function. In this paper, $p$-type norm is taken as cost functions, \emph{i.e.}, $C(\mathbf{z}_1, \mathbf{z}_2)=\|\mathbf{z}_1-\mathbf{z}_2 \|_{p}$. In addition, $\Pi(\mathbb{P}_1,\mathbb{P}_2)$ represents the set of all probability distributions of $\mathbf{z}_1 \in \mathbb{R}^d$ and $\mathbf{z}_2 \in \mathbb{R}^d$ with marginal distribution $\mathbb{P}_1$ and $\mathbb{P}_2$.
\label{definition-Wasserstein-distance}
\end{definition}

Especially, when $\mathbb{P}_1$ and $\mathbb{P}_2$ are normal distributions, type-2 Wasserstein distance of $\mathbb{P}_1$ and $\mathbb{P}_2$ can be calculated with only the first two order moments.
\begin{proposition}
(\cite{givens1984class}, Proposition 7 ). If $p = 2$ and $\alpha=2$, the type-2 Wasserstein distance of two normal distribution $\mathbb{P}_1 = \mathcal{N}(\bm{\mu}_1,\mathbf{\Sigma}_1)$ and $\mathbb{P}_2 = \mathcal{N}(\bm{\mu}_2,\mathbf{\Sigma}_2)$, where $\bm{\mu}_1,\bm{\mu}_2 \in \mathbb{R}^d$ are mean vectors and $\mathbf{\Sigma}_1, \mathbf{\Sigma}_2 \in \mathbb{S}^d_+$ are covariance matrices, then
    \begin{equation}
    	\begin{aligned}
        W_{2}(\mathbb{P}_1,\mathbb{P}_2)=\sqrt{\parallel \bm{\mu}_1 - \bm{\mu}_2\parallel_2^2 +\text{Tr}\left[\mathbf{\Sigma}_1+\mathbf{\Sigma}_2 - 2\left(\mathbf{\Sigma}_2^{\frac{1}{2}} \mathbf{\Sigma}_1\mathbf{\Sigma}_2^{\frac{1}{2}}\right)^{\frac{1}{2}} \right]}.
        \end{aligned}
        \label{Wasserstein-distance-normal}
    \end{equation}
\label{propostition-Wasserstein-distance-normal}
\end{proposition}



\section{Wasserstein Robust Graph Learning With Gaussian Prior Assumptions}
\label{sec:formulation-gaussian}
\subsection{Reformulation as Convex Optimization}
In this section, we assume that all distributions in uncertainty set $\mathbb{P}$ are normal distributions. It is known that smooth signals over a graph are governed by normal distribution, one case of which is mentioned in \cite{kalofolias2016learn}. Thanks to the nice properties of normal distributions, if we choose cost function as type-2 norm function and $\alpha =2$, the Wasserstein distance of two normal distributions has closed form as displayed in (\ref{Wasserstein-distance-normal}). One advantage of equation (\ref{Wasserstein-distance-normal}) is that the type-2 Wasserstein distance is explicitly available while first two order moments of Gaussian distribution are provided. Based on (\ref{Wasserstein-distance-normal}), the uncertainty sets containing only normal distribution can then be constructed as follows.

Suppose $\mathbb{P} \sim \mathcal{N}(\bm{\mu},\mathbf{\Sigma})$ and empirical distribution $\mathbb{P}_{n} \sim \mathcal{N}(\bm{\mu}_n,\mathbf{\Sigma}_n)$, then uncertainty set centered at nominal distribution $\mathcal{P} = \left\{ \mathbb{P} :  W(\mathbb{P}, \mathbb{P}_{n}) \leq \epsilon \right \}$ is equivalent to
\begin{equation}
    \begin{aligned}
   \mathcal{P}_{G} = \bigg\{(\bm{\mu},\mathbf{\Sigma}): \sqrt{\parallel \bm{\mu}_n - \bm{\mu}\parallel_2^2 +\text{Tr}\left[\mathbf{\Sigma}+\mathbf{\Sigma}_n - 2\left(\mathbf{\Sigma}_n^{\frac{1}{2}} \mathbf{\Sigma}\mathbf{\Sigma}_n^{\frac{1}{2}}\right)^{\frac{1}{2}} \right]}
    \leq \epsilon, \mathbf{\Sigma} \in \mathbb{S}_{+}^{d}\bigg\}.
    \end{aligned}
    \label{uncertainty set-gaussian}
\end{equation}
Therefore, if $p=2$ and $\alpha =2$, (\ref{basic infsup formulation}) can be written as
	\begin{equation}
        \begin{aligned}
            \underset{\mathbf{L}\in \mathcal{L}}{\text{inf}} R(\mathbf{L})  = \underset{\mathbf{L}\in \mathcal{L}}{\text{inf}}\, \underset{\mathbb{P}\in \mathcal{P}_G}{\text{sup}}\, \mathbb{E}_\mathbb{P} \left[\mathbf{x}^{\text{T}}\mathbf{L}\mathbf{x}+\eta\parallel\mathbf{L} \parallel_{F}^2\right]
        \end{aligned}
        \label{gaussian formulation basic form}
    \end{equation}
At a first glance, it is not tractable to solve the worst case $R(\mathbf{L})$ of $(\ref{gaussian formulation basic form})$ because of the expectation operator. For this reason, we move to the dual form of worst case risk $R(\mathbf{L})$ such that we may  get rid of the expectation operator.

\begin{lemma}
If $\alpha = 2$, $p=2$, for any $\gamma \geq  0$, we have
	\begin{equation}
        R(\mathbf{L})
        =\underset{\gamma\geq  0}{\text{inf}}\,\, 
        h(\gamma, \mathbf{L})+
        \gamma\left[\epsilon^2 - \text{Tr}(\mathbf{\Sigma}_n)\right]
        +\eta\parallel\mathbf{L} \parallel_{F}^2,
        \label{gaussian formulation dual form}
    \end{equation}
where
\begin{equation}
h(\gamma, \mathbf{L})=  \underset{\bm{\mu}, \mathbf{\Sigma} \succeq \mathbf{0}}{\text{sup}}\,\bigg\{\text{Tr}\big{(}\mathbf{\Sigma}(\mathbf{L} - \gamma \mathbf{I})\big{)} +\text{Tr}(\mathbf{L}\bm{\mu}\bm{\mu}^{\text{T}})-\gamma\parallel \bm{\mu} - \bm{\mu}_n\parallel_2^2 + 2\gamma\text{Tr} \left(\sqrt{\mathbf{\Sigma}_n^{\frac{1}{2}} \mathbf{\Sigma} \mathbf{\Sigma}_n^{\frac{1}{2}}}\right)  \bigg\}.
\end{equation}
\label{proposition-gaussian formulation dual form}
\end{lemma}

 \begin{IEEEproof}
 The proof of Lemma \ref{proposition-gaussian formulation dual form} is straightforward. Firstly notice that
 \begin{align}
 R(\mathbf{L}) &= \underset{\mathbb{P}\in\mathcal{P}_G}{\text{sup}}\,\,\mathbb{E}_\mathbb{P}\left[\mathbf{x}^{\text{T}}\mathbf{L}\mathbf{x}+\eta\parallel\mathbf{L} \parallel_{F}^2\right] \\
 &= \underset{\mathbb{P}\in\mathcal{P}_G}{\text{sup}}\,\,\text{Tr}(\mathbf{L}\mathbb{E}_\mathbb{P}[\mathbf{x}\mathbf{x}^{\text{T}}])+\eta\parallel\mathbf{L} \parallel_{F}^2 \\
 &= \underset{\bm{\mu},\mathbf{\Sigma}\succeq 0}{\text{sup}}\,\,\text{Tr}(\mathbf{L}(\mathbf{\Sigma} + \bm{\mu}\bm{\mu}^{\text{T}}))+\eta\parallel\mathbf{L} \parallel_{{F}}^2.
 \end{align}
The worst case risk can then be expressed as:
	\begin{equation}
        \begin{aligned}
        &\underset{\bm{\mu},\mathbf{\Sigma}\succeq 0}{\text{sup}}\,\,\text{Tr}(\mathbf{L}(\mathbf{\Sigma} + \bm{\mu}\bm{\mu}^{\text{T}}))+\eta\parallel\mathbf{L} \parallel_{{F}}^2\\
        &\text{s.t.}\,\, \sqrt{\parallel \bm{\mu}_n - \bm{\mu}\parallel^2 +\text{Tr}\left[\mathbf{\Sigma}+\mathbf{\Sigma}_n - 2\left(\mathbf{\Sigma}_n^{\frac{1}{2}} \mathbf{\Sigma}\mathbf{\Sigma}_n^{\frac{1}{2}}\right)^{\frac{1}{2}} \right]} \leq \epsilon
        \end{aligned}
        \label{proof - gaussian formulation dual form}
    \end{equation}
By dualizing the constraint of (\ref{proof - gaussian formulation dual form}), we can obtain 	\begin{equation}
        \begin{aligned}
        R(\mathbf{L}) &= 
                \underset{\bm{\mu},\mathbf{\Sigma}\succeq 0}{\text{sup}} \underset{\gamma \geq 0}{\text{inf}}\,\,\text{Tr}(\mathbf{L}(\mathbf{\Sigma} + \bm{\mu}\bm{\mu}^{\text{T}}))+\eta\parallel\mathbf{L} \parallel_{{F}}^2 + \gamma \left(\epsilon^2 -  \parallel \bm{\mu}_n - \bm{\mu}\parallel^2 -\text{Tr}\bigg{(}\mathbf{\Sigma}+\mathbf{\Sigma}_n - 2\big{(}\mathbf{\Sigma}_n^{\frac{1}{2}} \mathbf{\Sigma}\mathbf{\Sigma}_n^{\frac{1}{2}}\big{)}^{\frac{1}{2}} \bigg{)}\right)\\
                & = \underset{\gamma \geq 0}{\text{inf}}\,\,\gamma\left[\epsilon^2 - \text{Tr}(\mathbf{\Sigma}_n)\right]
                +\eta\parallel\mathbf{L} \parallel_{F}^2 + \underset{\bm{\mu},\mathbf{\Sigma}\succeq 0}{\text{sup}}\,\, \text{Tr}\big{(}\mathbf{\Sigma}(\mathbf{L} - \gamma \mathbf{I})\big{)} +\text{Tr}(\mathbf{L}\bm{\mu}\bm{\mu}^{\text{T}})-\gamma\parallel \bm{\mu} - \bm{\mu}_n\parallel_2^2 + 2\gamma\text{Tr} \left(\sqrt{\mathbf{\Sigma}_n^{\frac{1}{2}} \mathbf{\Sigma} \mathbf{\Sigma}_n^{\frac{1}{2}}}\right)  \\
                &=\underset{\gamma\geq  0}{\text{inf}}\,\, 
        h(\gamma, \mathbf{L})+
        \gamma\left[\epsilon^2 - \text{Tr}(\mathbf{\Sigma}_n)\right]
        +\eta\parallel\mathbf{L} \parallel_{F}^2.
        \end{aligned}
        \label{proof - gaussian formulation dual form-1}
    \end{equation}
The second equality holds because of strong duality \cite{nguyen2018distributionally}. With (\ref{proof - gaussian formulation dual form-1}), we can reach the conclusion of Lemma \ref{proposition-gaussian formulation dual form}.
 \end{IEEEproof}

Although the expectation operation is avoided by using duality, the above problem (\ref{gaussian formulation dual form}) is still hard to handle due to the inner supreme problem. Therefore, we further calculate the closed form of the inner supreme problem and then convert the original problem into a tractable convex optimization form.
\begin{theorem}
If $\alpha =2, p=2$, we have
	\begin{equation}
        \begin{aligned}
       \inf_{\mathbf{L} \in \mathcal{L} }R(\mathbf{L})=  &\underset{\mathbf{L}\in \mathcal{L},\gamma \mathbf{I} \succ \mathbf{L}}{\text{inf}} \gamma(\epsilon^2 - \text{Tr}(\mathbf{\Sigma}_x) ) + \gamma^2\text{Tr}((\gamma\mathbf{I} - \mathbf{L})^{-1} \mathbf{\Sigma}_x)+\eta\parallel\mathbf{L} \parallel_{{F}}^2,
        \end{aligned}
        \label{gaussian reformulation}
    \end{equation}
where $\mathbf{\Sigma}_x \triangleq \mathbf{\Sigma}_n +\bm{\mu}_n\bm{\mu}_n^{\text{T}}$.
\label{therorem-gaussian reformulation}
\end{theorem}

The proof of Theorem \ref{therorem-gaussian reformulation} is placed in Appendix \ref{sec:appendix-1}. With Theorem \ref{therorem-gaussian reformulation}, we may get rid of expectation operator and the inf-sup problem. Then it is feasible to solve (\ref{gaussian reformulation}) to obtain a desired graph.

\subsection{Solving the Convex Optimization}
For convenience, we define 
\begin{equation}\label{eqn:def-1}
g(\gamma, \mathbf{L}) \triangleq \gamma(\epsilon^2 - \text{Tr}(\mathbf{\Sigma}_x) ) + \gamma^2\text{Tr}((\gamma\mathbf{I} - \mathbf{L})^{-1} \mathbf{\Sigma}_x)+\eta\parallel\mathbf{L} \parallel_{F}^2,\ \ \ (\gamma>\lambda_{max}).    
\end{equation}
where $\lambda_{max}$ is the largest eigenvalue of $\mathbf{L}$. We add this constraint because of the constraint $\gamma \mathbf{I} \succ \mathbf{L}$ in (\ref{gaussian reformulation}). In order to solve (\ref{gaussian reformulation}), we propose that $g(\gamma, \mathbf{L})$ is convex as the following theorem \ref{covexity} states.

\begin{theorem}\label{covexity}
For $\gamma>\lambda_{max}$ and $\mathbf{L} \in \mathcal{L}$, problem \eqref{gaussian reformulation} is convex with $\gamma$ and $\mathbf{L}$.
\end{theorem}

Before we prove Theorem \ref{covexity}, some important facts, which is essential to our proof, are provided

\begin{lemma}
(Facts 7.4.8 in \cite{bernstein2009matrix}) For any $\mathbf{A}, \mathbf{B} \in \mathbb{R}^{d\times d}$ and $\mathbf{C} \in \mathbb{S}^d$, then 
	\begin{equation}
        \begin{aligned}
        \text{Tr}(\mathbf{A}\mathbf{C}\mathbf{B}\mathbf{C}) = \text{vec}(\mathbf{C})^{\text{T}}(\mathbf{B}\otimes\mathbf{A}^{\text{T}})\text{vec}(\mathbf{C})
        \label{equation krocker}
        \end{aligned}
    \end{equation}
\label{lemma-krocker}
\end{lemma}

\begin{lemma}
(Proposition 7.1.7 in \cite{bernstein2009matrix}) If $\mathbf{A}, \mathbf{B} \in \mathbb{S}^d_{+}$, then 
    \begin{equation}
        \begin{aligned}
        (\mathbf{A}\otimes\mathbf{B})^{-1} = \mathbf{A}^{-1}\otimes\mathbf{B}^{-1} 
        \label{equation krocker-inverse}
        \end{aligned}
    \end{equation}
\label{lemma-krocker-inverse}
\end{lemma}

\begin{IEEEproof}
The proof is similar with Proposition 4.3 in \cite{nguyen2018distributionally}. Since $g(\gamma, \mathbf{L})$ is a multivariable function with $\gamma$ and $\mathbf{L}$ jointly, we calculate the Hessian matrix of $g(\gamma, \mathbf{L})$ and prove the positive definite of it. Through basic calculation, the Hessian matrix of $g(\gamma, \mathbf{L})$ is shown as follows.

\begin{equation}
    \begin{aligned}
    \mathbf{H} &= 
    \begin{bmatrix} \frac{\partial^2 g(\gamma, \mathbf{L})}{\partial \text{vec}(\mathbf{L})^2} & \frac{\partial^2 g(\gamma, \mathbf{L})}{\partial \gamma \partial \text{vec}(\mathbf{L}) } \\ \left(\frac{\partial^2 g(\gamma, \mathbf{L})}{\partial \gamma \partial \text{vec}(\mathbf{L}) } \right)^{\text{T}}& \frac{\partial^2 g(\gamma, \mathbf{L})}{\partial \gamma^2} \\ \end{bmatrix} \\
    &= \begin{bmatrix} 2\gamma^2\left( ((\gamma\mathbf{I} - \mathbf{L})^{-1}\mathbf{\Sigma}_x(\gamma\mathbf{I} - \mathbf{L})^{-1})\otimes(\gamma\mathbf{I} - \mathbf{L})^{-1} \right) + 2\eta\mathbf{I}_{d^2} &  2\gamma\text{vec}((\gamma\mathbf{I} - \mathbf{L})^{-1}(\mathbf{I} - \gamma(\gamma\mathbf{I} - \mathbf{L})^{-1})\mathbf{\Sigma}_x (\gamma\mathbf{I} - \mathbf{L})^{-1})  \\  2\gamma\text{vec}((\gamma\mathbf{I} - \mathbf{L})^{-1}(\mathbf{I} - \gamma(\gamma\mathbf{I} - \mathbf{L})^{-1})\mathbf{\Sigma}_x (\gamma\mathbf{I} - \mathbf{L})^{-1})^{\text{T}} & 2\text{Tr}\left(\mathbf{\Sigma}_x(\mathbf{I} - \gamma(\gamma\mathbf{I} - \mathbf{L})^{-1})^2(\gamma\mathbf{I} - \mathbf{L})^{-1} \right) \end{bmatrix}
    \label{hessian of g}
    \end{aligned}
\end{equation}
where $\otimes$ is Kronecker product and $\mathbf{I}_{d^2}$ means identity matrix with $d^2$ dimensions.

Next, we are aimed to prove the positive definite of $\mathbf{H}$. Firstly, since $\gamma > \lambda_{max}$ and $\mathbf{\Sigma}_x\succ 0$, then $ \frac{\partial^2 g(\gamma, \mathbf{L})}{\partial \text{vec}(\mathbf{L})^2}\succ 0$. secondly, we need to  calculate the Schur complement of $ \frac{\partial^2 g(\gamma, \mathbf{L})}{\partial \text{vec}(\mathbf{L})^2}$ in $\mathbf{H}$ 

\begin{equation}
    \begin{aligned}
    S &= \frac{\partial^2 g(\gamma, \mathbf{L})}{\partial \gamma^2} -
    \left(\frac{\partial^2 g(\gamma, \mathbf{L})}{\partial \gamma \partial \text{vec}(\mathbf{L}) } \right)^{\text{T}}  \bigg{(}2\gamma^2\left( ((\gamma\mathbf{I} - \mathbf{L})^{-1}\mathbf{\Sigma}_x(\gamma\mathbf{I} - \mathbf{L})^{-1})\otimes(\gamma\mathbf{I} - \mathbf{L})^{-1} \right) + 2\eta\mathbf{I}_{d^2}\bigg{)}^{-1} \frac{\partial^2 g(\gamma, \mathbf{L})}{\partial \gamma \partial \text{vec}(\mathbf{L}) } \\
    & > \frac{\partial^2 g(\gamma, \mathbf{L})}{\partial \gamma^2} -
    \left(\frac{\partial^2 g(\gamma, \mathbf{L})}{\partial \gamma \partial \text{vec}(\mathbf{L}) } \right)^{\text{T}}  \bigg{(}2\gamma^2\left( ((\gamma\mathbf{I} - \mathbf{L})^{-1}\mathbf{\Sigma}_x(\gamma\mathbf{I} - \mathbf{L})^{-1})\otimes(\gamma\mathbf{I} - \mathbf{L})^{-1} \right) \bigg{)}^{-1} \frac{\partial^2 g(\gamma, \mathbf{L})}{\partial \gamma \partial \text{vec}(\mathbf{L}) }\\
    & =  \frac{\partial^2 g(\gamma, \mathbf{L})}{\partial \gamma^2} -\frac{1}{2\gamma^2}
    \left(\frac{\partial^2 g(\gamma, \mathbf{L})}{\partial \gamma \partial \text{vec}(\mathbf{L}) } \right)^{\text{T}}      \left( (\gamma\mathbf{I} - \mathbf{L})\mathbf{\Sigma}_x^{-1}(\gamma\mathbf{I} - \mathbf{L})\otimes(\gamma\mathbf{I} - \mathbf{L}) \right)  \frac{\partial^2 g(\gamma, \mathbf{L})}{\partial \gamma \partial \text{vec}(\mathbf{L}) }\\
    & = \frac{\partial^2 g(\gamma, \mathbf{L})}{\partial \gamma^2} -  \frac{1}{2\gamma^2}\text{Tr}\left((\gamma\mathbf{I} - \mathbf{L})\frac{\partial^2 g(\gamma, \mathbf{L})}{\partial \gamma \partial \mathbf{L}}(\gamma\mathbf{I} - \mathbf{L})\mathbf{\Sigma}_x^{-1}(\gamma\mathbf{I} - \mathbf{L}) \frac{\partial^2 g(\gamma, \mathbf{L})}{\partial \gamma \partial \mathbf{L}}
    \right)\\
    & = 0
    \label{equation schur-compliment}
    \end{aligned}
\end{equation}
The inequality holds because $2\eta\mathbf{I}_{d^2} \succ 0$. The second and the third equality holds due to lemma \ref{lemma-krocker-inverse} and lemma \ref{lemma-krocker} respectively. We can observe from (\ref{equation schur-compliment}) that the Schur complement of $ \frac{\partial^2 g(\gamma, \mathbf{L})}{\partial \text{vec}(\mathbf{L})^2}$ in $\mathbf{H}$ is greater than 0. Thus, $\mathbf{H}$ is a positive definite matrix. Finally, the constraints for the domain of the function $g(\gamma,\mathbf{L})$, \emph {i.e}., $\gamma > \lambda_{max}$ and $\mathbf{L} \in \mathcal{L}$ do not affect the convexity of the function $g(\gamma,\mathbf{L})$. Thus we reach the conclusion of theorem \ref{covexity}.
\end{IEEEproof}

\begin{figure}[t] 
    \centering
       \includegraphics[width=0.48\linewidth]{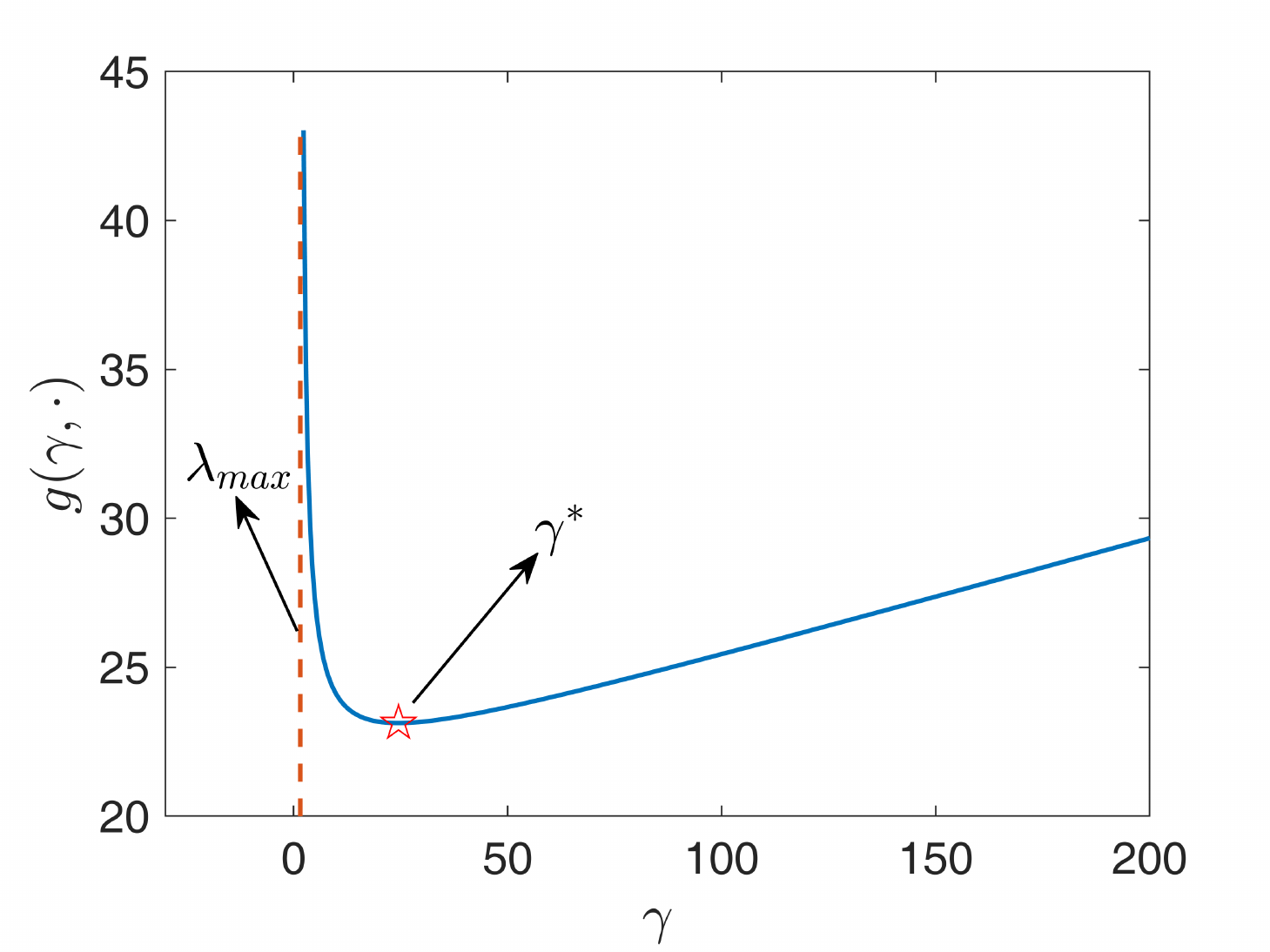}
    	\caption{Trend of $g(\gamma,\cdot)$ with $\gamma$ for a given $\mathbf{L}$}
    	\label{fig-g-gamma}
\end{figure}

Theorem \ref{covexity} implies that there exists an unique  solution $(\gamma^*, \mathbf{L}^*)$ for problem (\ref{gaussian reformulation}). Thus it is natural  to solve \eqref{gaussian reformulation} by using a Newton-type method. Albeit feasible, we give up this method owing to the complexity and numerical stability. Instead, we adopt a block coordinate descent method to solve (\ref{gaussian reformulation}). First of all, we fix $\mathbf{L}$ and update $\gamma$. The corresponding function $g(\gamma,\cdot)$ is  nonlinear and convex. To better illustrate the characteristics of the function $g(\gamma,\cdot)$, for a given $\mathbf{L}$, Figure \ref{fig-g-gamma} plots the trend of $g(\gamma,\cdot)$ with $\gamma$. It is clear that $g(\gamma, \cdot)$ is a convex function, and will explode to infinity quickly when $\gamma$ approaches $\lambda_{max}$. On the other hand, $g(\gamma, \cdot)$ will also asymptotically linearly toward infinity when $\gamma$ toward infinity.

In order to find the minimizer $\gamma^*$ of $g(\gamma, \cdot)$, we calculate the first-order partial derivative of $g(\gamma, \mathbf{L})$ on $\gamma$, 
	\begin{equation}
        \begin{aligned}
        g_{\gamma}(\gamma,\mathbf{L}) = \frac{\partial g(\gamma, \mathbf{L}) }{\partial \gamma}
        & = \epsilon^2 - \text{Tr}(\mathbf{\Sigma}_x) + 2\gamma\text{Tr}((\gamma\mathbf{I} - \mathbf{L})^{-1} \mathbf{\Sigma}_x) -\gamma^2\text{Tr}((\gamma\mathbf{I}-\mathbf{L})^{-1}\mathbf{\Sigma}_x(\gamma\mathbf{I}-\mathbf{L})^{-1})\\
        &= \epsilon^2- \text{Tr}\left((\mathbf{I} - \gamma(\gamma \mathbf{I} - \mathbf{L})^{-1})^2 \mathbf{\Sigma}_x\right).\label{derivative with gamma}
        \end{aligned}
    \end{equation}
What we need to do right now is to find $\gamma^*$ such that $g_{\gamma}(\gamma^*,\mathbf{L}) = 0$. Usually one might apply  Newton method to directly calculate $\gamma^*$. However, experiments show that if the initial value is not selected properly, the numerical results will diverge frequently. To address this issue, we resort bisection method to find $\gamma^*$. Actually,  it locates between the biggest eigenvalue  $\lambda_{max}$ of graph Laplacian operator and the infinity as Figure \ref{fig-g-gamma} depicts, henceforth it is feasible to use bisection method to find $\gamma^*$.
The flow of bisection method is shown in Algorithm 1.

\begin{remark}
We may represent \eqref{eqn:def-1} as 
\begin{equation}\label{eqn:def-1-app}
g(\gamma, \mathbf{L}) \triangleq \gamma(\epsilon^2 - \text{Tr}(\mathbf{\Sigma}_x) ) + \gamma^2\cdot\frac{Q_{d-1}(\gamma)}{\gamma\cdot P_{d-1}(\gamma)}    +\eta\parallel\mathbf{L} \parallel_{F}^2,\ \ \ (\gamma>\lambda_{max}), 
\end{equation}
in which both $P_{d-1}(\gamma)$ and $Q_{d-1}(\gamma)$ are polynomials of order $d-1$ and 
\begin{equation}
 P_{d-1}(\gamma)=(\gamma-\lambda_2)(\gamma-\lambda_3)\cdots (\gamma-\lambda_{max}),
\end{equation}
which leads to the asymptotically linear behavior while $\gamma\to +\infty$. Notice that $0=\lambda_1<=\lambda_2\le \lambda_{max}$ are eigenvalues of Laplacian $\mathbf{L}$. 
\label{remark-asymptotic}
\end{remark}

\begin{algorithm}[t] 
\caption{Update $\gamma$ using bisection search} 
\label{alg:bisection} 
\begin{algorithmic}[1] 
\REQUIRE ~~\\ 
Tolerance $error$;\\
Uncertainty set size $\epsilon$\\
Parameters of search area $a,b$\\
Empirical mean vector $\bm{\mu}_n$;\\
Empirical covariance matrix $\mathbf{\Sigma}_n$\\
Graph Laplacian of last iteration $\mathbf{L}$;\\
\ENSURE ~~\\ 
The minimizer $\gamma^*$;
\STATE Calculate $\lambda_{\text{max}}$ of $\mathbf{L}$;\\
Set initial upper bound $ub = a\times \lambda_{\text{max}}$;\\
Set initial lower bound $lb = \lambda_{\text{max}} + b$;\\
\STATE Calculate initial middle value $mid =({ub+lb})/{2}$;
\label{ code:cal mid }
\STATE Calulate $g_{\gamma}(lb, \mathbf{L})$, $g_{\gamma}(ub, \mathbf{L})$ and $g_{\gamma}(mid,\mathbf{L})$ using (\ref{derivative with gamma});
\STATE If $g_{\gamma}(lb,\mathbf{L})g_{\gamma}(mid,\mathbf{L}) < 0$:\\
        {\,\,\,\,\,\,}Update $ub$, $ub = mid$;\\
       If $g_{\gamma}(ub,\mathbf{L})g_{\gamma}(mid,\mathbf{L}) < 0$:\\
         {\,\,\,\,\,\,}Update $lb$, $lb = mid$;\\
\STATE If $ub - lb > error$:\\
        {\,\,\,\,\,\,}Go back to step 2;\\
        Else:\\
        {\,\,\,\,\,\,}$\gamma^* = ({ub+lb})/{2}$
        
\RETURN $\gamma^*$ 
\end{algorithmic}
\end{algorithm}

Secondly, we fix $\gamma$ and update $\mathbf{L}$, and \eqref{gaussian reformulation} boils down to the following problem.
\begin{equation}
    \begin{aligned}
     \underset{\mathbf{L}}{\text{inf}}\,\,\,\, & 
     \gamma^2\text{Tr}(\mathbf{\Sigma}_x(\gamma\mathbf{I}-\mathbf{L})^{-1}) + \eta\parallel \mathbf{L}\parallel_F^2,\\
     \text{s.t.}\,\, \,\, &{l}_{ij} = {l}_{ji} \leq 0,\,\, i \neq j,\\
     &  \mathbf{L}\mathbf{1} = \mathbf{0},\\
     & \text{Tr}(\mathbf{L}) = d.
    \end{aligned}
    \label{reformulation-update L}
\end{equation}

Note that, except $\mathbf{L} \in \mathcal{L}$, we add an extra constraint $\text{Tr}(\mathbf{L}) = d$, which is meant to avoid trivial solutions \cite{dong2016learning}. As lemma \ref{covexity} states, it is a convex optimization problem. We can apply interior point method to solve it and a variety of convex optimization package, such as CVX in MATLAB, can be adpoted. However, in interior point method, we need to solve a Newton type equation to obtain update directions, which is time consuming especially for order $d^2$ variables. Inspired by \cite{kumar2020unified}, we exploit a projection gradient descent (PGD) method to solve this problem. Since the symmetry of $\mathbf{L}$ and the $\mathbf{L}\mathbf{1} = \mathbf{0}$ constraint, the degrees of freedom of $\mathbf{L}$ is $\frac{d(d-1)}{2}$. Hence, we introduce a linear operator $\mathcal{T}$, which is defined in detail in Appendix \ref{sec:appendix-3}, that can convert a non-negative vector $\mathbf{v} \in \mathbb{R}^{{d(d-1)}/{2}}_+$ into a Laplacian matrix. With  $\mathcal{T}$, we can rewrite (\ref{reformulation-update L}) as
\begin{equation}
    \begin{aligned}
    \underset{\mathbf{v}\geq 0 }{\text{inf}}\,\,\,\, r(\mathbf{v}) =  \underset{\mathbf{v}\geq 0 }{\text{inf}}\,\,\,\, &\gamma^2\text{Tr}(\mathbf{\Sigma}_x(\gamma\mathbf{I} - \mathcal{T}\mathbf{v})^{-1}) + \eta \parallel \mathcal{T}\mathbf{v}\parallel_{\text{F}}^2 +\beta (\text{Tr}(\mathcal{T}\mathbf{v})-d)^2.
    \end{aligned}
    \label{rewrite with linear operator 2}
\end{equation}
The last term of (\ref{rewrite with linear operator 2}) is a relaxation of the constraint $\text{Tr}(\mathcal{T}\mathbf{v}) = d$. With this relaxation, solving a convex problem with equality constraint is avoided. Indeed, (\ref{rewrite with linear operator 2}) is a convex problem of $\mathbf{v}$ without structural constraints, and hence we can apply projection gradient descent (PGD) method to solve it. Notice that the derivative of $r(\mathbf{v})$ is
\begin{equation}
    \begin{aligned}
    \nabla r(\mathbf{v}) = \gamma^2\mathcal{T}^*(\mathbf{(\gamma\mathbf{I} - \mathcal{T}\mathbf{v})^{-2}\Sigma}_x) + 2 \eta \mathcal{T}^*\mathcal{T}\mathbf{v} + 2\beta(\text{Tr}(\mathcal{T}\mathbf{v})-d)\mathcal{T}^*\mathbf{I}.
    \end{aligned}
    \label{deriviate of r}
\end{equation}
Then we may update $\mathbf{v}$ by using (\ref{deriviate of r}),
\begin{equation}
    \begin{aligned}
        \mathbf{v}_{k+1} = (\mathbf{v}_{k} - s \nabla r(\mathbf{v}_k))_{+},
    \end{aligned}
    \label{update of L in gaussian}
\end{equation}
where $(\cdot)_+ \triangleq \text{max}(\cdot,0)$, and $s$ is the step size that is determined using backtrack line search method\cite{nocedal2006numerical}. Finally, we can reach the complete flow of our proposed block coordinate descent algorithm, which is shown in Algorithm 3.

Note that $g(\gamma,\mathbf{L})$ is jointly convex with $\gamma$ and $\mathbf{L}$, hence there exists a unique global minimum $(\gamma^*, \mathbf{L}^*)$. Specifically, in the domain of $g(\gamma, \mathbf{L})$, all terms are actually differentiable, and we can reach the conclusion that $g(\gamma, \mathbf{L})$ is regular at each coordination-wise minimum point $\gamma^*$ and $\mathbf{L}^*$ based on Lemma 3.1 in \cite{tseng2001convergence}. Therefore, the minimum point $\gamma^*$ and $\mathbf{L}^*$ for each coordinate are the stationary points of $g(\gamma,\mathbf{L})$, which is also the global minimum point for a convex function. Based on this, the results obtained by iteratively updating $\gamma$ and $\mathbf{L}$ will converge monotonically to the global minimum $\gamma^*, \mathbf{L}^*$ finally.

\begin{algorithm}[t] 
\caption{Update $\mathbf{L}$ using PGD} 
\label{alg:PGD-Gaussian} 
\begin{algorithmic}[1] 
\REQUIRE ~~\\ 
$\mathbf{\Sigma}_x$,\,$\eta$,\,$\beta$,\, $\gamma^*$ of last iteration, \,tolerance $error$, \,\\max iteration $maxIter$;\\
\ENSURE ~~\\ 
The learned graph $\mathbf{L}^*$;\\
\STATE Initialize $iter = 0$, $\mathbf{v}>0$;\\
\STATE Calculate update direction using (\ref{deriviate of r});
\STATE Calculate step size $s$ using line search method;
\STATE Update $\mathbf{v}_{iter+1}$ using (\ref{update of L in gaussian});
\STATE $iter = iter + 1;$
\STATE If $\parallel \mathbf{v}_{iter} -  \mathbf{v}_{iter-1}\parallel_{\text{2}} \geq error$ and $iter<maxIter$:\\
{\,\,\,\,\,\,} Go back to Step 2;\\
Else:\\
    {\,\,\,\,\,\,} Go to return Step;
\RETURN $\mathbf{L}^* = \mathcal{T}\mathbf{v}$
\end{algorithmic}
\end{algorithm}

\begin{algorithm}[t] 
\caption{ Graph learning in gaussian scenario} 
\label{alg:Coordinate descent} 
\begin{algorithmic}[1] 
\REQUIRE ~~\\ 
Tolerance $error$;\\
Uncertainty set size $\epsilon$;\\
Max iteration $maxIter$;\\
Empirical mean vector $\bm{\mu}_n$;\\
Empirical covariance matrix $\mathbf{\Sigma}_n$;\\
\ENSURE ~~\\ 
The learned graph $\mathbf{L}^*$;\\
\STATE 
Initialize $iter = 0$;\\
Initialize any $\mathbf{L}_0 \in \mathcal{L}_c$;\\
Set $\mathbf{\Sigma}_x = \mathbf{\Sigma}_n + \bm{\mu}_n\bm{\mu}_n^{\text{T}}$;\\
\STATE Update $\gamma_{iter+1}$ using Algorithm 1 with $\mathbf{L}_{iter}$;
\STATE Update $\mathbf{L}_{iter+1}$ by solving $(\ref{reformulation-update L})$ with $\gamma_{iter+1}$;\\
\STATE $iter = iter + 1;$
\STATE If $\parallel \mathbf{L}_{iter} -  \mathbf{L}_{iter-1}\parallel_{\text{F}} \geq error$ and $iter<maxIter$:\\
{\,\,\,\,\,\,} Go back to Step 2;\\
Else:\\
    {\,\,\,\,\,\,} Go to return Step;

\RETURN $\mathbf{L}_{iter}$ 
\end{algorithmic}
\end{algorithm}

\subsection{Computation Complexity}
In the step of updating $\gamma$, one of the time-consuming step is calculating $\lambda_{max}$. The common method to calculate eigenvalue is Cholesky decomposition, which requires order $\mathcal{O}(d^3/3)$ flops. For large-scale graph, we can apply Lanczos \cite{Lanczos1950AnIM} to calculate the $\lambda_{max}$ of a symmetric matrix, such as $\mathbf{L}$. Moreover, we do not need to calculate $\lambda_{max}$ at each iteration. As the number of iterations increases, the difference between each $\lambda_{max}$ will be small, hence the previously calculated $\lambda_{max}$ can be used as the 
lower bound of bisection. On the other hand, when we calculate (\ref{derivative with gamma}), the most time consuming is inverse operator whose complexity is $\mathcal{O}(d^3)$. We can also apply spectral perturbation method to avoid inverse operation at each iteration. Based on above analysis, in updating $\gamma$ step, the costs is $p_1d^3$, $p_1$ is the number of iterations of bisection method to converge.

In the step of updating $\mathbf{L}$, the cost of both $\mathcal{T}$ and $\mathcal{T}^*$ is order $\mathcal{O}(d^2)$. When calculating derivative, the complexity is $\mathcal{O}(d^3)$. Then updating $\mathbf{L}$ costs order $\mathcal{O}(p_2d^3)$ flops, where $p_2$ is the number of iterations of PGD method to converge.

Based on the above analysis, the overall procedure costs order $p_3\text{max}\{p_1d^3,p_2d^3\}$ flops, and $p_3$ is the number of iterations of Algorithm 3.

\section{Wasserstein Robust Graph Learning Without Prior Assumption}
\label{sec:formulation-general}
\subsection{Reformulation as Convex Optimization}
In this section, we consider the case when prior information of the distribution of signals is not assumed, \emph{i.e.}, uncertainty set might contain any distribution whose Wasserstein distance from empirical distribution smaller than $\epsilon$. Furthermore, distributions are not required to be the same type of Wasserstein distance. In this setting, the uncertainty set can be defined as
\begin{equation}
\begin{aligned}
\mathcal{P} = \left\{\mathbb{P}: W_{\alpha} (\mathbb{P},\mathbb{P}_n)  \leq \epsilon,\,\, \text{for all distributions } \mathbb{P}  \right\}
\end{aligned}
\label{uncertainty set general}
\end{equation}

Revisit problem (\ref{basic infsup formulation}),  the general formulation can then be specified as 

\begin{equation}
\begin{aligned}
\underset{\mathbf{L}\in \mathcal{L}}{\text{inf}} R(\mathbf{L})  =
 \underset{\mathbf{L}\in \mathcal{L}}{\text{inf}}\, \underset{\mathbb{P}\in \mathcal{P}}{\text{sup} }\, &\mathbb{E}_\mathbb{P} \left[\mathbf{x}^{\text{T}}\mathbf{L}\mathbf{x}+\eta\parallel\mathbf{L} \parallel_{{F}}^2\right]
\end{aligned}
\label{basic infsup formulation in general}
\end{equation}

We firstly focus on the worst case risk $R(\mathbf{L})$. Since no assumption is made on distributions, it is not feasible to use the first two order moments to replace expectation operator just as the previous section does. Obviously, expectation operator in (\ref{wrost case formulation}) is defined on the probability of infinite dimensions, and it is difficult to solve such an optimization problems directly. For the same reason as we present in Lemma \ref{proposition-gaussian formulation dual form}, the following dual form of (\ref{basic infsup formulation in general}) is obtained by using the definition of $W_{\alpha}(\mathbb{P},\mathbb{P}_n)$,
\begin{lemma}
For any $\gamma > 0$, the dual form of the worst case risk in (\ref{basic infsup formulation in general}) can be written as:
\begin{equation}
    \begin{aligned}
     R(\mathbf{L}) 
     & =  \underset{\mathbb{P}\in \mathcal{P}}{\text{sup}}\, \mathbb{E}_\mathbb{P} \left[\mathbf{x}^{\text{T}}\mathbf{L}\mathbf{x}+\eta\parallel\mathbf{L} \parallel_{{F}}^2\right ]\\ 
    & = \underset{\gamma\geq  0}{\text{inf}}\,\, \bigg\{ \gamma \epsilon^{\alpha}+\eta\parallel\mathbf{L} \parallel_{\text{F}}^2  + \frac{1}{N} \sum_{i=1}^N  \underset{\mathbf{x} \in {\mathbb{R}}^d} {\text{sup}} \left\{\text{Tr}(\mathbf{x}^{\text{T}}\mathbf{L}\mathbf{x})-\gamma \parallel \mathbf{x} - \mathbf{x}_i\parallel_p^{\alpha} \right\}   \bigg\},
    \end{aligned}
    \label{formulation dual form in general}
\end{equation}
\label{proposition-dual-form-general}
\end{lemma}
The proof of lemma \ref{proposition-dual-form-general} is derived from Corollary 2 in \cite{gao2016distributionally}, and the minor change is that cost function is chosen as $p$-type norm in our formulation. With the dual form, we can easily reach corollary \ref{corollary dual form Q}. 
\begin{corollary}
The dual form of (\ref{basic infsup formulation in general}) can be further reformulated into the following form
\begin{equation}
    \begin{aligned}
     R(\mathbf{L}) 
     &= \underset{\mathbb{Q}\in\mathcal{Q}}{\text{sup}}\,\,\mathbb{E}_\mathbb{Q}[\text{Tr} (\mathbf{L} \mathbf{\Theta})+\eta\parallel\mathbf{L} \parallel_{\text{F}}^2] \\ 
    & = \underset{\gamma\geq  0}{\text{inf}}\,\, \bigg\{ \gamma \epsilon^{\alpha}+\eta\parallel\mathbf{L} \parallel_{\text{F}}^2 + \frac{1}{N} \sum_{i=1}^N  \underset{\mathbf{U} \in {\mathbb{S}}^d} {\text{sup}} \left\{\text{Tr}(\mathbf{L}\mathbf{U})-\gamma \parallel \textbf{vec}(\mathbf{U}) -  \textbf{vec}(\mathbf{\Theta}_i)\parallel_p^{\alpha} \right\}   \bigg\},
    \end{aligned}
    \label{dual form Q}
\end{equation}
\label{corollary dual form Q}
where $\mathbf{\Theta} \triangleq \mathbf{x}\mathbf{x}^{\text{T}}$, $\mathbf{\Theta}_i = \mathbf{x}_i\mathbf{x}_i^{\text{T}}$. $\mathbb{Q}$ is the distribution of $\mathbf{\Theta}$ and $\mathcal{Q}$ is the set of $\mathbb{Q}$ induced by $\mathcal{P}$.
\end{corollary}
Here the first equality holds because 
\begin{equation}
 \begin{aligned}
    R(\mathbf{L})
    &= \underset{\mathbb{P}\in \mathcal{P}}{\text{sup}}\, \mathbb{E}_\mathbb{P} [\mathbf{x}^{\text{T}}\mathbf{L}\mathbf{x}+\eta\parallel\mathbf{L} \parallel_{\text{F}}^2] \\
    & = \underset{\mathbb{P}\in \mathcal{P}}{\text{sup}}\, \mathbb{E}_\mathbb{P} [\text{Tr} (\mathbf{L}\mathbf{x}\mathbf{x}^{\text{T}})+\eta\parallel\mathbf{L} \parallel_{\text{F}}^2] \\
    &= \underset{\mathbb{Q}\in\mathcal{Q}}{\text{sup}}\,\,\mathbb{E}_\mathbb{Q}[\text{Tr} (\mathbf{L} \mathbf{\Theta})+\eta\parallel\mathbf{L} \parallel_{\text{F}}^2],
\end{aligned}
\label{proof-Q-form}
\end{equation}
and the second equality of (\ref{dual form Q}) holds due to the duality presented in lemma \ref{proposition-dual-form-general}. In (\ref{dual form Q}), we replace the origin variable $\mathbf{x}$ with $\mathbf{\Theta}$ because a closed form of the inner supreme problem of ($\ref{dual form Q}$) can be calculated with variable $\mathbf{\Theta}$. Specifically, with variable $\mathbf{\Theta}$, we can prove that (\ref{basic infsup formulation}) in general scenario is equivalent to a convex problem using its dual form.
\begin{theorem}
The solution of (\ref{basic infsup formulation}) in general scenario is equivalent to the following problem.
    \begin{equation}
        \begin{aligned}
        &\underset{\mathbf{L}\in \mathcal{L}}{\text{inf}}\, \underset{\mathbb{P}\in \mathcal{P}}{\text{sup}}\, \mathbb{E}_\mathbb{P} [\mathbf{x}^{\text{T}}\mathbf{L}\mathbf{x}+\eta\parallel\mathbf{L} \parallel_{\text{F}}^2] \\
        = &\underset{\mathbf{L}\in \mathcal{L}}{\text{inf}}\, \text{Tr}(\mathbf{L}\mathbf{\Theta}_n) +\eta\parallel\mathbf{L} \parallel_{\text{F}}^2+ \epsilon \parallel \mathbf{vec}(\mathbf{L}) \parallel_q
        \end{aligned}
        \label{reformulation-regularize}
    \end{equation}
where $\mathbf{\Theta}_n = \frac{\mathbf{\Theta}_1 +\mathbf{\Theta}_2 + .... \mathbf{\Theta}_N}{N} = \frac{\mathbf{x}_1{\mathbf{x}_1^{\text{T}}+\mathbf{x}_2}{\mathbf{x}_2^{\text{T}}+...+\mathbf{x}_N}{\mathbf{x}_N^{\text{T}}}}{N}$, and $\frac{1}{p}+\frac{1}{q} = 1$.
\label{theorem-regularize}
\end{theorem}

The detailed proof of Theorem \ref{theorem-regularize} is illustrated in Appendix \ref{sec:appendix-2}. Note that the right side of (\ref{reformulation-regularize}) is actually an SAA problem plus a regularization term. It is an interesting conclusion because robustness can be seen as a regularization term and the size of uncertainty set $\epsilon$ controls the weight of regularization term as well as the level of robustness. Therefore, our proposed framework provides an interpretation for the relationship between the size (radius) of uncertainty sets and robustness level. 

\subsection{Solving the Convex Optimization}
To solve (\ref{reformulation-regularize}), we also add an extra constraint $\text{Tr}(\mathbf{L}) = d$ as the previous section does for the same reason and the complete problem is shown as follows.
	\begin{equation}
        \begin{aligned}
          \underset{\mathbf{L}}{\text{inf}}\,\,\,\, &\text{Tr}(\mathbf{L}\mathbf{\Theta}_n) + \eta \parallel \mathbf{L}\parallel_{\text{F}}^2+\epsilon \parallel \mathbf{vec}(\mathbf{L}) \parallel_q\\
         \text{s.t.}\,\, \,\, &\l_{ij} = l_{ji} \leq 0,\,\, i \neq j\\
         &  \mathbf{L}\mathbf{1} = \mathbf{0}\\
         & \text{Tr}(\mathbf{L}) = d
        \end{aligned}
        \label{algorithm-reformulation-regularize}
    \end{equation}
If $q \geq 1$, which is tantamount to $p \geq 1$, the object function and constraints are all convex. We can solve this convex problem by using projection gradient descent just as above-mentioned. With the linear operator $\mathcal{T}$, the origin problem (\ref{algorithm-reformulation-regularize}) can be rewritten as:
\begin{equation}
    \begin{aligned}
   & \underset{\mathbf{v}\geq 0 }{\text{inf}}\,\, m(\mathbf{v}) =  \text{Tr}(\mathbf{\Theta}_n\mathcal{T}\mathbf{v}) + \eta \parallel \mathcal{T}\mathbf{v}\parallel_{\text{F}}^2 + \epsilon \parallel \mathbf{vec}(\mathcal{T}\mathbf{v}) \parallel_q +\beta (\text{Tr}(\mathcal{T}\mathbf{v})-d)^2
    \end{aligned}
    \label{rewrite with linear operator}
\end{equation}
The derivative of $m(\mathbf{v})$ is 
	\begin{equation}
        \begin{aligned}
        \nabla m(\mathbf{v}) = &\mathcal{T}^*\mathbf{\Theta}_n + 2 \eta \mathcal{T}^*\mathcal{T}\mathbf{v} + 2\beta(\text{Tr}(\mathcal{T}\mathbf{v})-d)\mathcal{T}^*\mathbf{I} + \epsilon\parallel \mathbf{vec}(\mathcal{T} \mathbf{v})\parallel_q^{1-q}\mathcal{T}^*(\mathcal{T}\mathbf{w})^{.(q-1)},
        \end{aligned}
        \label{deriviate of m}
    \end{equation}
where $(\cdot)^{.(q-1)}$ means an element-wise operation. We can then update $\mathbf{v}$ using PGD
	\begin{equation}
        \begin{aligned}
            \mathbf{v}_{k+1} = (\mathbf{v}_{k} - s \nabla m(\mathbf{v_k}))_{+},
        \end{aligned}
        \label{pgd update}
    \end{equation}
where $(\cdot)_+ \triangleq \text{max}(\cdot,0)$, and $s$ is the step size that is determined using backtrack line search method\cite{nocedal2006numerical}. The complete algorithm flow is shown in Algorithm 4. The convergence of our algorithms can be guaranteed since it is a convex problem.

\begin{algorithm}[t] 
\caption{ Graph learning in general scenario} 
\label{alg:PGD} 
\begin{algorithmic}[1] 
\REQUIRE ~~\\ 
$\mathbf{\Theta}_n$,\,$\eta$,\,$\beta$,\, uncertainty set size $\epsilon$, \,tolerance $error$, \,\\max iteration $maxIter$;\\
\ENSURE ~~\\ 
The learned graph $\mathbf{L}^*$;\\
\STATE Initialize $iter = 0$, $\mathbf{v}>0$;\\
\STATE Calculate update direction using (\ref{deriviate of m});
\STATE Calculate step size $s$ using line search method;
\STATE Update $\mathbf{v}_{iter+1}$ using (\ref{pgd update});
\STATE $iter = iter + 1;$
\STATE If $\parallel \mathbf{v}_{iter} -  \mathbf{v}_{iter-1}\parallel_{\text{2}} \geq error$ and $iter<maxIter$:\\
{\,\,\,\,\,\,} Go back to Step 2;\\
Else:\\
    {\,\,\,\,\,\,} Go to return Step;
\RETURN $\mathbf{L}^* = \mathcal{T}\mathbf{v}$
\end{algorithmic}
\end{algorithm}

\subsection{Computation Complexity}
In Algorithm 4,  both $\mathcal{T}$ and $\mathcal{T}^*$ cost order $\mathcal{O}(d^2)$ flops. In addition, since $(\cdot)^{.(q-1)}$ is a elementwise operator, the cost is $\mathcal{O}(d^2)$. Therefore, the overall procedure costs order $p_4d^2$ flops, where $p_4$ is the number of iterations of PGD method to converge. When the scale of graph is large, which implies $d$ is a large value, the speed of PGD will be superior to that of interior point method.

\section{Reformulation as Semi-Definite Programming (SDP)}
\label{sec:formulation-sdp}
In both Gaussian and general scenario, we convert the intractable (\ref{basic infsup formulation}) into a problem that can be solved easily. In this section, we will prove that, if $\alpha =2$ and $p =2$ for both scenarios, (\ref{basic infsup formulation}) is equivalent to a SDP problem. However, we will also illustrate that it is impractical to solve such SDP problems and that is why we do not learn a robust graph in this way.

We first display the SDP form of Gaussian scenario, which is shown in proposition \ref{proposition-gaussian formulation SDP}
\begin{proposition}
If $\alpha = 2$, $p=2$, in Gaussian scenario, the following SDP problem and problem (\ref{basic infsup formulation}) have the same optimal value.
	\begin{equation}
        \begin{aligned}
        \underset{\mathbf{L}\in \mathcal{L}, \gamma \mathbf{I} \succ \mathbf{L},z>=0, \mathbf{Z}\in \mathbb{S}_{+}^d }{\text{inf}} &\gamma(\epsilon^2 - \text{Tr}(\mathbf{\Sigma}_x))+\eta\parallel\mathbf{L} \parallel_{\mathbf{F}}^2 +z +\text{Tr}(\mathbf{Z})\\
        \text{s.t.}\,\,         
        &\begin{bmatrix} \gamma\mathbf{I} - \mathbf{L} & \gamma \mathbf{\mu}_n \\ \gamma         \mathbf{\mu}_n^{\text{T}} & z \\ \end{bmatrix}\succeq 0, \\
        &\begin{bmatrix} \gamma\mathbf{I} - \mathbf{L} & \gamma \mathbf{\Sigma}_n^{\frac{1}{2}} \\ \gamma         \mathbf{\Sigma}_n^{\frac{1}{2}} & \mathbf{Z}  \\ \end{bmatrix}\succeq 0
        \end{aligned}
        \label{gaussian formulation SDP}
    \end{equation}

\label{proposition-gaussian formulation SDP}
\end{proposition}

\begin{IEEEproof}
 From Theorem \ref{therorem-gaussian reformulation}, the origin inf-sup problem has been proven to be tantamount to an inf-problem (\ref{gaussian reformulation}). The remaining proof is similar with Corollary 2.9 in \cite{nguyen2018distributionally}. Specifically, in (\ref{gaussian reformulation}), we focus our attention on the nonlinear term $ \phi(\gamma, \mathbf{L}) \triangleq \gamma^2\text{Tr}((\gamma\mathbf{I}-\mathbf{L})^{-1}\mathbf{\Sigma}_x) = \gamma^2\text{Tr}((\gamma\mathbf{I}-\mathbf{L})^{-1}\mathbf{\Sigma}_n) + \gamma^2\text{Tr}((\gamma\mathbf{I}-\mathbf{L})^{-1}\mathbf{\mu}_n\mathbf{\mu}_n^{\text{T}})$. The domain of $\phi(\gamma, \mathbf{L})$ is $\{(\gamma, \mathbf{L}) : \gamma\mathbf{I}\succ \mathbf{L}, \mathbf{L} \in \mathcal{L}\}$.In fact, $\phi(\gamma,\mathbf{L})$ is a matrix fractional function described in \cite{boyd2004convex} and has the following reformulation \cite{nguyen2018distributionally}:
    
    \begin{equation}
     \begin{aligned}
        \phi(\gamma, \mathbf{L}) 
        &= \underset{t}{\text{inf}}\left\{t \,:\, \,\gamma\mathbf{I}\succ \mathbf{L},\, \gamma^2\text{Tr}((\gamma\mathbf{I}-\mathbf{L})^{-1}\mathbf{\Sigma}_n) + \gamma^2\text{Tr}((\gamma\mathbf{I}-\mathbf{L})^{-1}\mathbf{\mu}_n\mathbf{\mu}_n^{\text{T}}) \leq t \right\}\\
        & =\underset{t, \mathbf{Z},z}{\text{inf}}\left\{t \,:\,\, \gamma\mathbf{I}\succ \mathbf{L},\,\mathbf{Z} \succeq 
        \gamma^2\mathbf{\Sigma}_n^{\frac{1}{2}}(\gamma\mathbf{I}-\mathbf{L})^{-1}\mathbf{\Sigma}_n^{\frac{1}{2}},\, z \geq \gamma^2\mathbf{\mu}_n^{\text{T}}(\gamma\mathbf{I}-\mathbf{L})^{-1}\mathbf{\mu}_n,\, \text{Tr}(\mathbf{Z}) + z \leq t \right\}\\
        & =\underset{\mathbf{Z},z}{\text{inf}}\,\left\{\text{Tr}(\mathbf{Z}) + z \,:\,\, \gamma\mathbf{I}\succ \mathbf{L},\, \begin{bmatrix} \gamma\mathbf{I} - \mathbf{L} & \gamma \mathbf{\mu}_n \\ \gamma         \mathbf{\mu}_n^{\text{T}} & z \\ \end{bmatrix}\succeq 0,\, \begin{bmatrix} \gamma\mathbf{I} - \mathbf{L} & \gamma \mathbf{\Sigma}_n^{\frac{1}{2}} \\ \gamma         \mathbf{\Sigma}_n^{\frac{1}{2}} & Z  \\ \end{bmatrix}\succeq 0 \right\}\\
    \end{aligned}
    \label{proof-sdp-gaussian}
    \end{equation}
    Bring (\ref{proof-sdp-gaussian}) back to (\ref{gaussian reformulation}), we can finally reach the conclusion in proposition \ref{proposition-gaussian formulation SDP}.
\end{IEEEproof}

On the other hand, the SDP problem of general scenario is described in proposition  \ref{proposition-genegral formulation SDP}
\begin{proposition}
If $\alpha=2$, $p=2$, in general scenario, the following SDP problem and problem (\ref{basic infsup formulation}) have the same optimal value.
    \begin{equation}
     \begin{aligned}
        & \underset{\gamma > 0, z_i > 0, \mathbf{L} \in \mathcal{L} }{\text{inf}}\,\, \gamma\epsilon^2 +\eta\parallel\mathbf{L} \parallel_{\text{F}}^2 +  \frac{1}{N}\sum_{i=1}^N z_i,\\
        & \text{s.t.} 
        \begin{bmatrix} \gamma\mathbf{I} - \mathbf{L} & \gamma \mathbf{x}_i \\ \gamma         \mathbf{x}_i^{\text{T}} & z_i + \gamma\parallel \mathbf{x}_i\parallel_2^2 \\ \end{bmatrix}\succeq 0, \,\,\,\,\text{for}\,\, i = 1, ..., N
    \end{aligned}
    \label{general sdp}
    \end{equation}
\label{proposition-genegral formulation SDP}
\end{proposition}

\begin{IEEEproof}
From proposition (\ref{proposition-dual-form-general}), we can obtain the dual form of (\ref{basic infsup formulation}). By introducing auxiliary variables $z_i$, the dual form of worst case equals to:
    \begin{equation}
     \begin{aligned}
        R(\mathbf{L})
        & = \underset{\gamma > 0, z_i > 0}{\text{inf}}\,\, \gamma\epsilon^2 +\eta\parallel\mathbf{L} \parallel_{\text{F}}^2 + \frac{1}{N}\sum_{i=1}^N z_i,\\
        & \text{s.t.} \,\, \underset{\mathbf{x}\in \mathbb{R}^d}{\text{sup}}
        \mathbf{x}^{\text{T}}\mathbf{L}\mathbf{x} - \gamma \parallel \mathbf{x} - \mathbf{x}_i\parallel_2^2\ \leq z_i \,\,\,\,\text{for}\,\, i = 1, ..., N \\
     & = \underset{\gamma > 0, z_i > 0}{\text{inf}}\,\,       \gamma\epsilon^2+\eta\parallel\mathbf{L} \parallel_{\text{F}}^2 + \frac{1}{N}\sum_{i=1}^N z_i,\\
        & \text{s.t.} \,\, \underset{\mathbf{x}\in \mathbb{R}^d}{\text{sup}}
        {\begin{bmatrix} -\mathbf{x} \\ 1\\ \end{bmatrix}}^{\text{T}}
        \begin{bmatrix}  \mathbf{L} - \gamma\mathbf{I}  & - \gamma \mathbf{x}_i \\ - \gamma         \mathbf{x}_i^{\text{T}} & - z_i - \gamma\parallel \mathbf{x}_i\parallel_2^2 \\ \end{bmatrix} 
        \begin{bmatrix} -\mathbf{x} \\ 1\\ \end{bmatrix} \leq 0,\,\,\,\, \text{for}\,\, i = 1, ..., N \\
     & =  \underset{\gamma > 0, z_i > 0}{\text{inf}}\,\,       \gamma\epsilon^2 +\eta\parallel\mathbf{L} \parallel_{\text{F}}^2+ \frac{1}{N}\sum_{i=1}^N z_i,\\
        & \text{s.t.} \,\, \begin{bmatrix} \gamma\mathbf{I}- \mathbf{L} & \gamma \mathbf{x}_i \\ \gamma         \mathbf{x}_i^{\text{T}} & z_i + \gamma\parallel \mathbf{x}_i\parallel_2^2 \\ \end{bmatrix} \succeq 0, \,\,\,\,\text{for}\,\, i = 1, ..., N \\
    \end{aligned}
    \label{proof of remark}
    \end{equation}
\end{IEEEproof}

We then insert  (\ref{proof of remark}) into (\ref{basic infsup formulation}), and finally reach (\ref{general sdp}). It seems feasible to solve (\ref{gaussian formulation SDP}) and  (\ref{general sdp}) to obtain the desired graph $\mathbf{L}$ because we can resort state-of-art interior point solvers for such SDP problems. However, we should mention that reformulating (\ref{basic infsup formulation}) to an SDP problem brings extra variables such as $z$ and $\mathbf{Z}$ in (\ref{gaussian formulation SDP}) and $z_i$ in (\ref{general sdp}). Another point is that more constraints are incurred in SDP problems. Therefore, the scale of SDP problems is significantly larger than our formulation in section \ref{sec:formulation-gaussian} and \ref{sec:formulation-general}. The impact of extra variables and constraints will be more intolerant if we use interior point method, which is the most common method exploited by existing optimization packages, to solve SDP problems. In fact, according to our experiments in the  running environment shown in section \ref{sec:Experiments}, if we use interior point to solve an SDP problem with scale exceeding 150 vertices, out of memory error will appear anyway. Therefore, just as \cite{nguyen2018distributionally} states, when the number of vertices becomes large, it is no longer practical to solve an SDP problem even for moderate values $d$. In addition, observe that, in (\ref{general sdp}), the number of constraints is actually equal to that of samples. To reduce uncertainty, a large number of samples may be collected leading to a huge problem scale of (\ref{general sdp}). Under this circumstance, it is impractical to solve such an SDP problem. In section \ref{sec:Experiments}, we will compare the runtime of our proposed formulation with SDP formulation of Gaussian scenario (SDP of general scenario is not practical to solve) to validate our analysis.

\section{Experiments}
\label{sec:Experiments}
In this section, we evaluate our proposed graph learning framework with both synthetic data and real world data. First of all, experimental settings are presented.
\subsection{Main Settings}
On the top of to-do list is generating groundtruth graphs for synthetic data. We first generate a similarity graph where edge weights represent the similarity of corresponding vertices.  The similarity is calculated using Gaussian radial basis function (RBF), namely $\text{exp}(-\text{dist}(i,j)^2/2\sigma^2)$, where $\text{dist}(i,j)$ represents the distance between vertex $i$ and vertex $j$ and $\sigma$ is the kernel function width. Edges will be retained only when the similarity between two vertices is greater than a certain threshold $\tau$. In addition, for pipeline task, we detect communities in an SBM graph, which is a stochastic graph consist of some clusters, because it is suitable for cluster tasks \cite{holland1983stochastic}. Four parameters of SBM graphs are required to be determined, namely the number of clusters, the number of nodes in each cluster, the probability of node connection between clusters and within clusters. 

Graph signals are generated from distribution ${\cal{N}}(0,\mathbf{L}^{\dag})$, and ${\dag}$ represents pseudo inverse. We generate graph signals in this way because it is one of the model generating smooth signals in \cite{kalofolias2016learn} and the smooth signals can be explained as a result of graph filtering. After obtaining the learned graph, we eliminate some unimportant edges, whose weights are less than $10^{-4}$, in order to make the learned graph more reasonable.

All algorithms are implemented by MATLAB and run on an Intel(R) CPU with 3.80GHz clock speed and 16GB of RAM.

\subsection{Synthetic data}
For constructing groundtruth graph, we randomly generate coordinates of 20 vertices in a unit square and calculate the similarity between these vertices using method above with parameter $\sigma = 0.5$ and $\tau = 0.7$. Laplacian matrix of the groundtruth graph are then calculated and we generate graph signals using above-mentioned method. For all experiments, white noises $w \sim {\cal{N}}(0,\sigma_w^2) $, $\sigma_w = 0.1$ are added to the generated data directly. 

Two metrics are adopted to evaluate the performance of the learned graph, which are Matthews correlation coefficient (MCC)\cite{powers2020evaluation} and difference of graphs (DOG) respectively. MCC is defined as:
\begin{equation}
    \begin{aligned}
        \text{MCC} = \frac{\text{TP}\cdot\text{TN} - \text{FP}\cdot \text{FN}}{\sqrt{(\text{TP+FP})(\text{TP+FN})(\text{TN+TP})(\text{TN+TN})}},
    \end{aligned}
    \label{MCC}
\end{equation}
where $TP$ (true positives) is the number of nonzero off-diagonal entries of groundtruth graph correctly identified in the learned graph while $FN$ (false negatives) is the number of those that are falsely identified as zeros. On the other hand, $TN$ (true negatives) is the number of zero off-diagonal entries that are correctly identified in the learned graph while $FP$ is the number of those misidentified as non-zeros. MCC is one of the most informative metrics for tasks of binary classification because it fuses all information of $TN,TP,FN$ and $FP$. In the context of graph learning, binary classification can be understood as whether one edge are learned or not. The value of MCC belongs to $[-1,+1]$, and $+1$ means that the learned graph exactly learns the existence of all edges in groundtruth graph while $-1$ can be interpreted as a total misidentification.

The other metric, DOG, is used to evaluate the difference between groundtruth and the learned graph, which is defined as 
\begin{equation}
    \begin{aligned}
      \text{DOG} = \frac{\parallel \mathbf{L}^* - \mathbf{L}_{\text{gt}}\parallel_{\text{F}}}{\parallel \mathbf{L}_{\text{gt}}\parallel_{\text{F}}},
    \end{aligned}
    \label{DOG}
\end{equation}
where $\mathbf{L}^*$ is the learned Laplacian matrix and $\mathbf{L}_{\text{gt}}$ is the groundtruth one.

The method of SAA model are taken as a baseline which is almost the same as \cite{dong2016learning} in section \ref{sec:problem-statement}. Parameter $\epsilon$ is selected according to metric MCC, that is, the $\epsilon$ corresponding to the largest MCC will be selected from a candidate set. To improve the reliability of the learning results, we run each experiment 20 times independently, and the final result is the average of all trials. 

\subsubsection{Impact of sample size $N$}
\begin{figure}[t] 
    \centering
	  \subfloat[]{
       \includegraphics[width=0.48\linewidth]{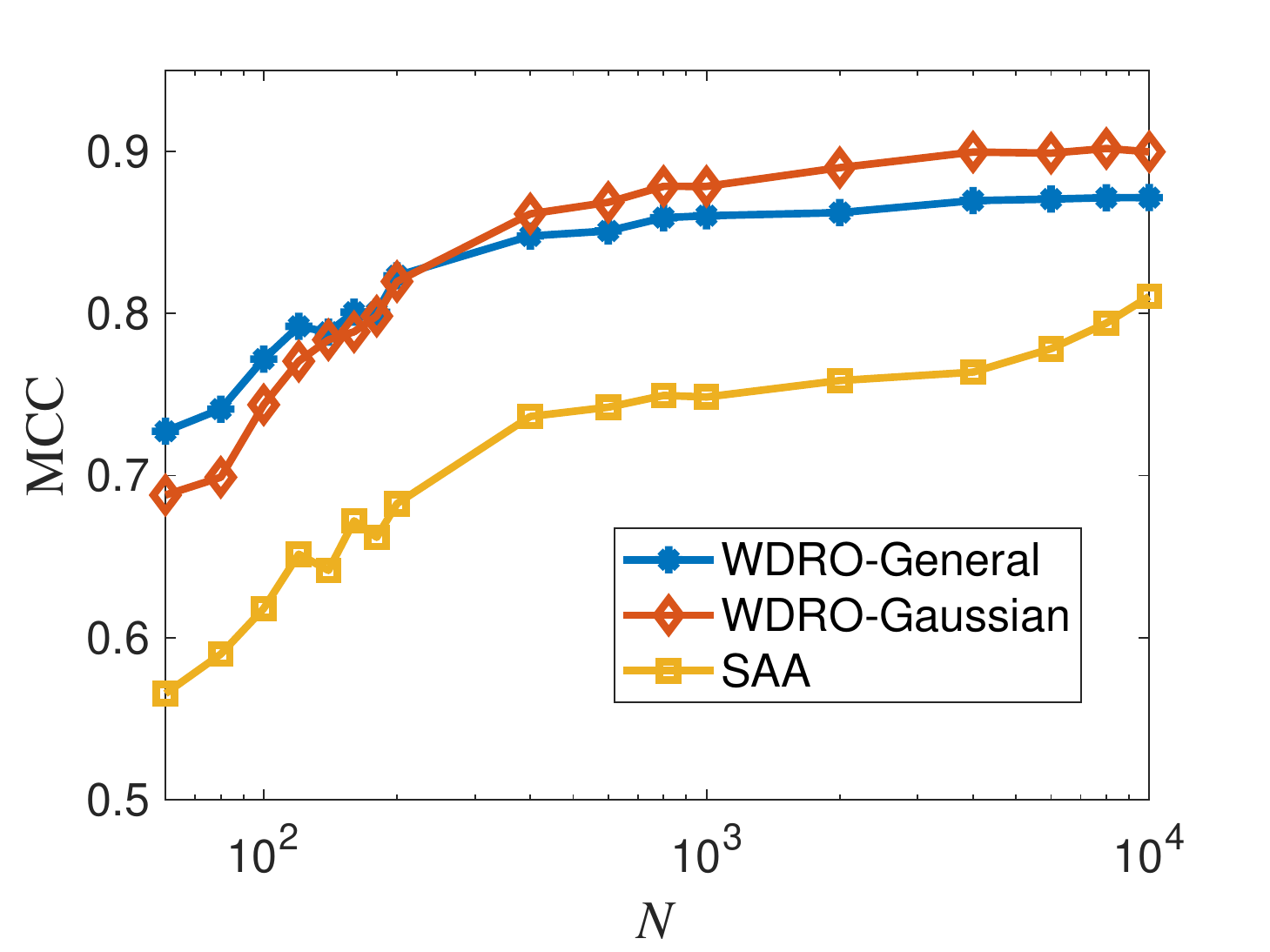}}
	  \quad
	  \subfloat[]{
        \includegraphics[width=0.48\linewidth]{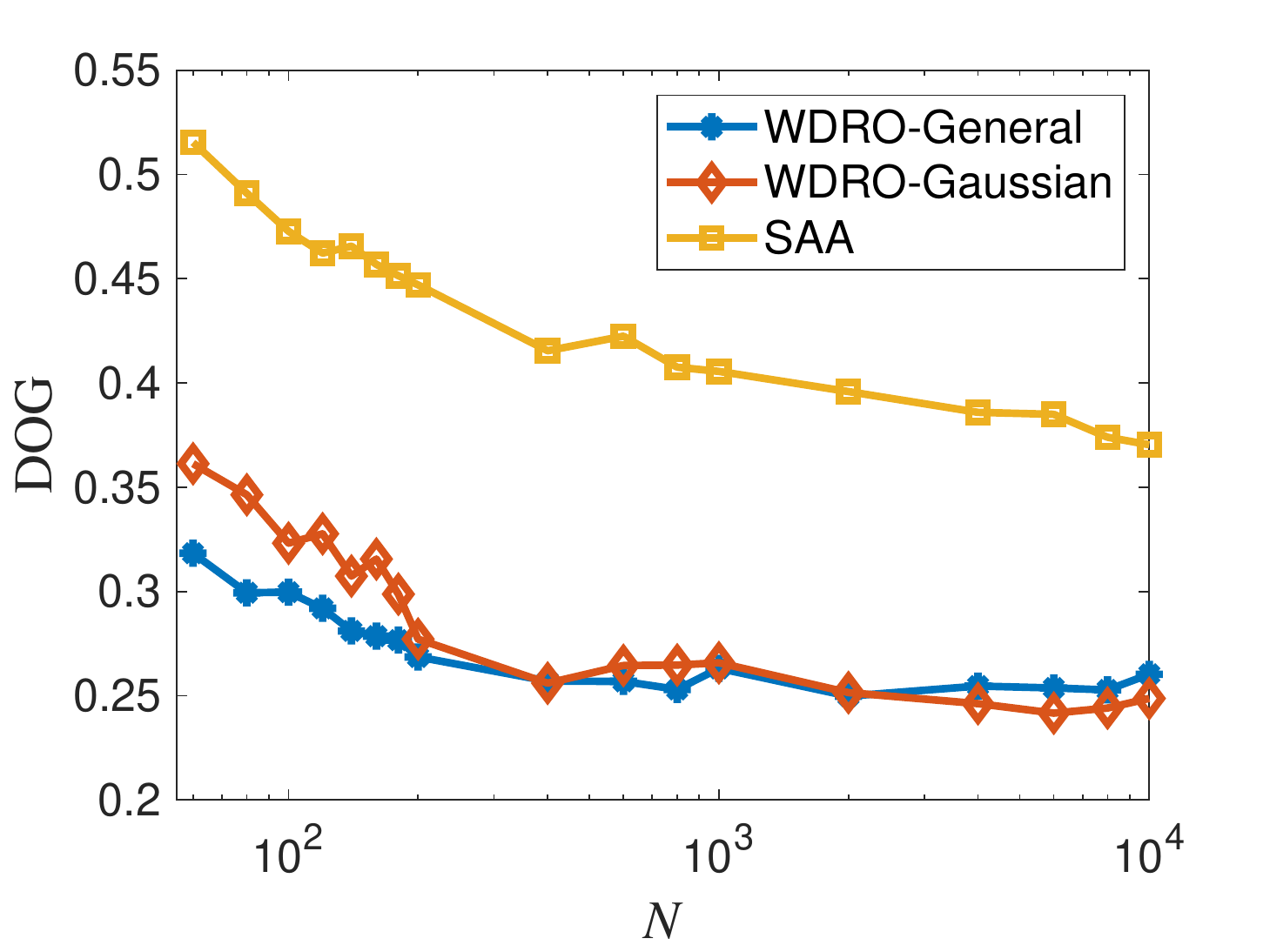}}
    	\caption{Performance of the learned graph with different sample size (a) MCC; (b) DOG }
    	\label{fig-N}
\end{figure}

We first study the uncertainty induced by sample size. Fig.\ref{fig-N} (a) depicts the relationship between MCC and the number of samples. When the number of samples are small, this indicates there is a large gap between nominal distribution and true distribution, which brings more uncertainty to samples. Therefore, the learned graph of SAA performs poorly in the case of small $N$. However, the performance of WDRO (both Gaussian and General scenario) is superior to that of SAA. As the number of samples increases, nominal distribution starts to approach the true distribution, leading to a decrease in sample uncertainty. This explains the fact that the performance of SAA increase as sample size. Additionally, the performance improvement of WDRO is small when the sample size is large due to the same reason mentioned before. Fig.\ref{fig-N} (b) shows the relationship between DOG and the number of samples and the result displays a similar trend with that of MCC and both of them illustrate the superiority of our framework.

\subsubsection{Impact of uncertainty set size $\epsilon$}
\begin{figure}[t] 
    \centering
	  \subfloat[]{
       \includegraphics[width=0.48\linewidth]{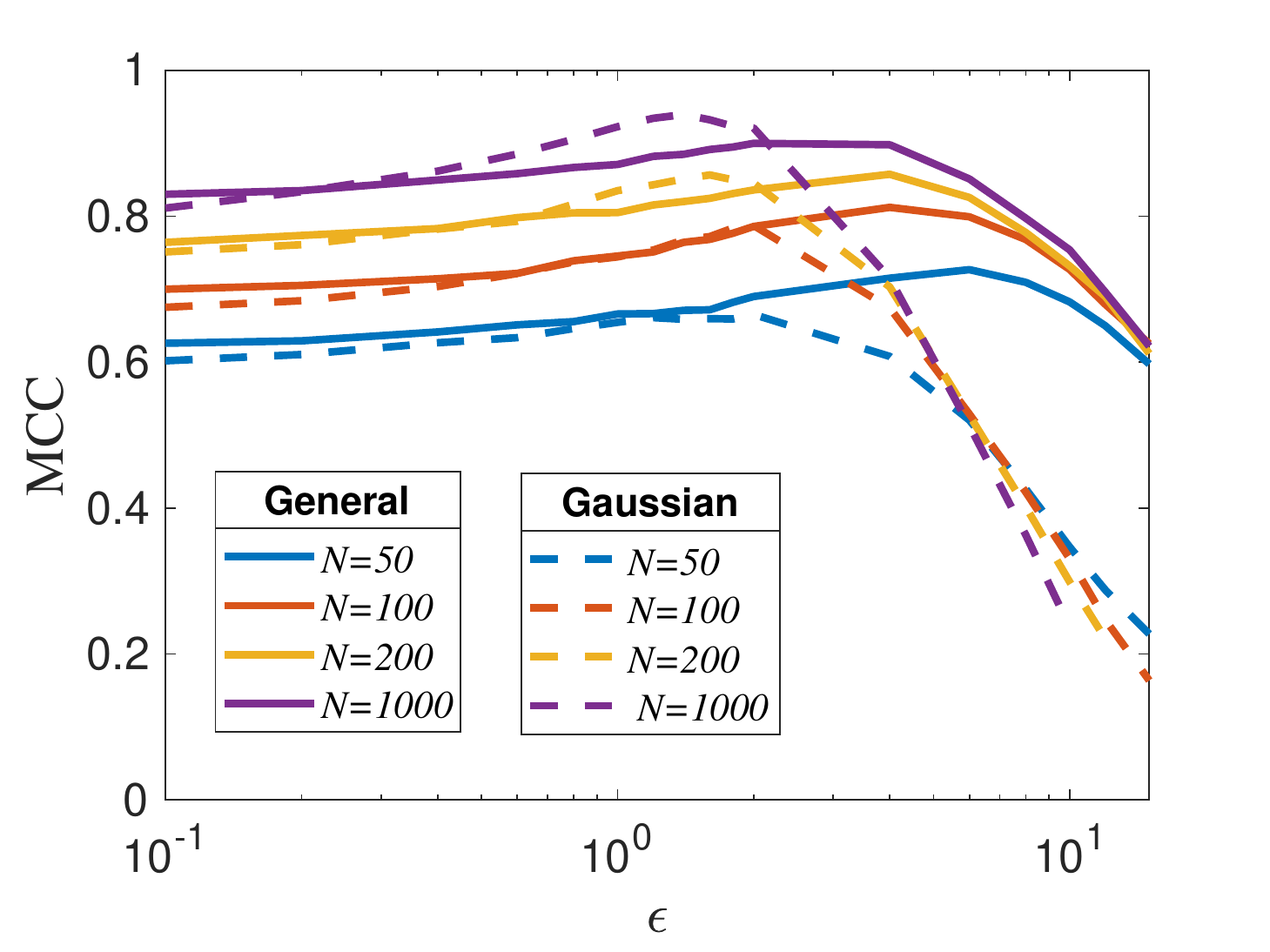}}
	  \quad
	  \subfloat[]{
        \includegraphics[width=0.48\linewidth]{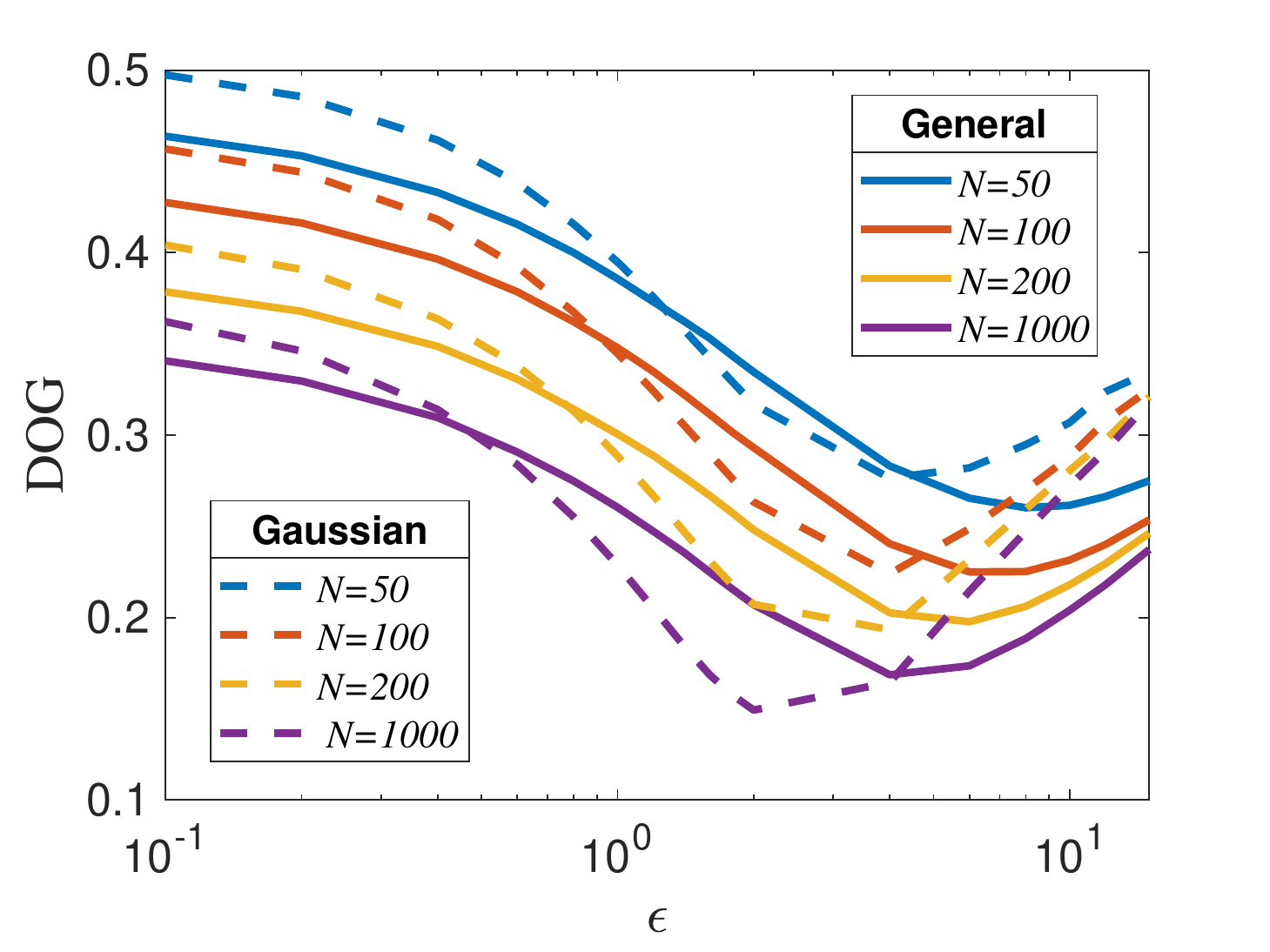}}
    	\caption{Performance of the learned graph with different uncertainty sizes (a) MCC; (b) DOG }
    	\label{fig-epsilon}
\end{figure}

We then check the impact of uncertainty set $\epsilon$ on the learned graph. We can see from \ref{fig-epsilon} (a) that, as $\epsilon$ increases, the value of MCC first increases and then drops down. This is due to that $\epsilon$ represents the size of uncertainty set and larger uncertainty set contains more distributions, which brings more robustness. For a certain level of uncertainty, when $\epsilon$ increases from 0, the level of robustness will start to approach to the one "best matching" the uncertainty level. However, if $\epsilon$ is too large, the result is not ideal because the uncertainty set may contain too many nuisance distributions far from the real one, causing the worst case risk is too conservative. The results vividly demonstrate the role of $\epsilon$ in our framework, and we need to make a trade-off between robustness levels. Furthermore, the best $\epsilon$, which is corresponding to the largest MCC or DOG value decreases as the number of sample increases. The reason for this trend is that increases in sample size reduces the uncertainty of sample distributions because the gap between nominal and real distributions is getting small as sample size increases. Therefore, it is suitable to set a lower robustness level for the case of large number of samples. Fig.\ref{fig-epsilon} (b) interprets the same trend from the perspective of DOG.

\subsubsection{$Reliability$}
To better understand the impact of $\epsilon$, we define a metric called certificate reliability as 
\begin{equation}
    \begin{aligned}
        \text{Reliability} = \mathbb{P}\{R(\mathbf{x}_t,\mathbf{L^*})<R^*\}, 
    \end{aligned}
    \label{raliability}
\end{equation}
where $R(\mathbf{x}_t,\mathbf{L^*})$ is the risk with respect to testing samples $\mathbf{x}_t$ and $\mathbf{L^*}$ and can be calculated as $\mathbf{x}_t^{\text{T}}\mathbf{L^*}\mathbf{x}_t +\eta \parallel \mathbf{L^*}\parallel_{\text{F}}^2$. Additionally, $R^*$ is the optimal value of (\ref{basic infsup formulation}) and is equivalent to the worst case $R(\mathbf{L}^*)$. Give some testing samples $\mathbf{x}_t$  and $\mathbf{L^*}$, reliability is tantamount to the empirical probability of that the risks of testing samples are smaller than $R^*$. On the other hand, reliability also represents the probability of whether uncertainty set contains the true distribution of signals. This can be explained that if uncertainty set is large enough to contain $\mathbb{P}_{\text{real}}$, the worst case risk $R(\mathbf{L}^*)$ of all distributions in uncertainty set must be larger than the risk of testing samples because testing samples and training samples are from the same distribution, that is, $\mathbb{P}_{\text{real}}$. 

\begin{figure}[t] 
    \centering
       \includegraphics[width=0.48\linewidth]{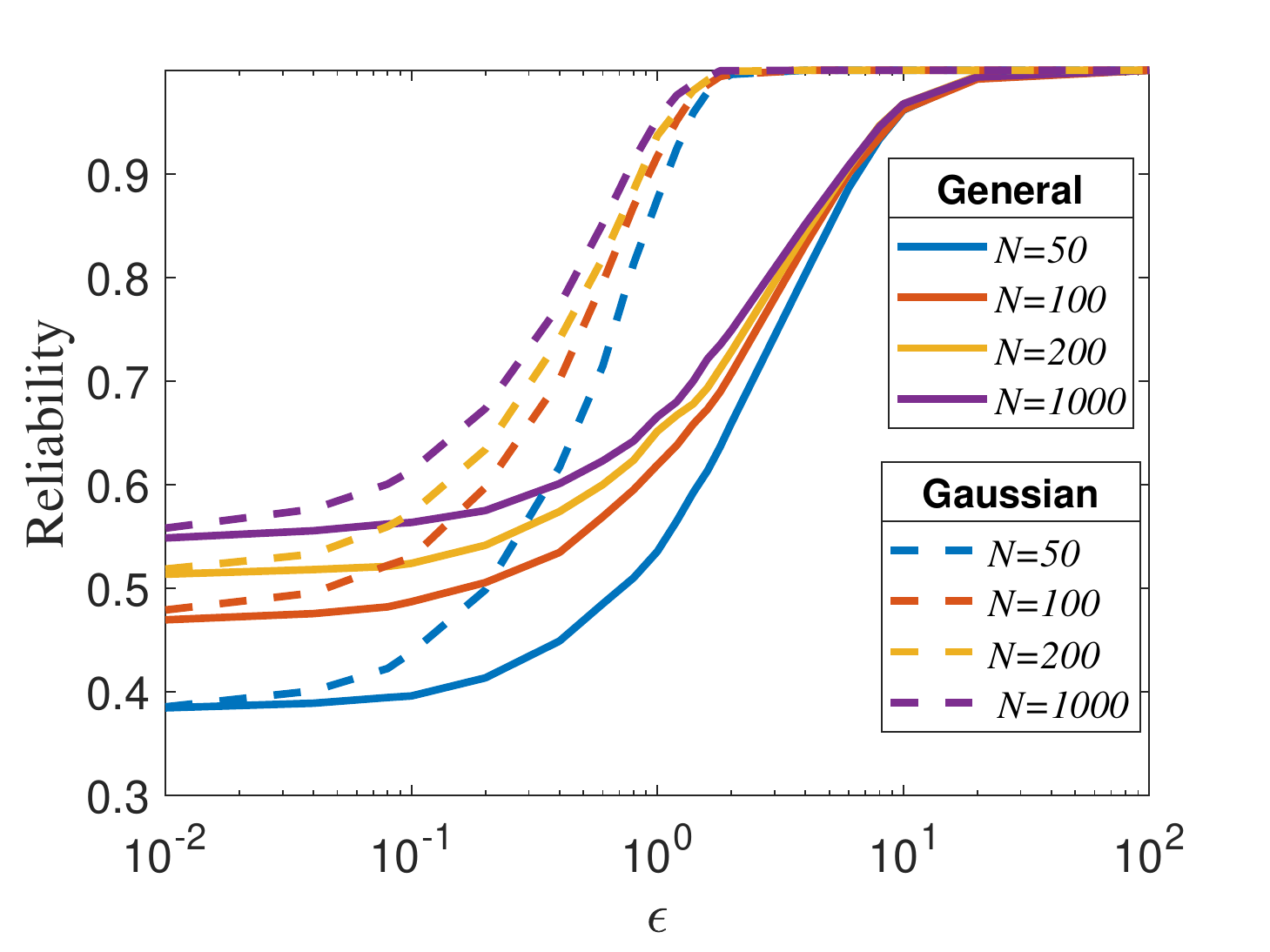}
    	\caption{Reliability values of different uncertainty set sizes}
    	\label{fig-reliability-epsilon}
\end{figure}

As shown is Fig.\ref{fig-reliability-epsilon}, for both general and Gaussian scenario, as $\epsilon$ increases, reliability approaches to 1. This trend makes sense since that for a larger $\epsilon$, uncertainty sets are more likely to contain $\mathbb{P}_{\text{real}}$.  Reliability reveals how $\epsilon$ controls robustness. Furthermore, from Fig.\ref{fig-reliability-epsilon}, we can conclude the fact that Gaussian scenario tends to need smaller $\epsilon$ for reliability reaching 1.

\subsubsection{Impact of noise level $\sigma_w$}
\begin{figure}[t] 
    \centering
	  \subfloat[]{
       \includegraphics[width=0.48\linewidth]{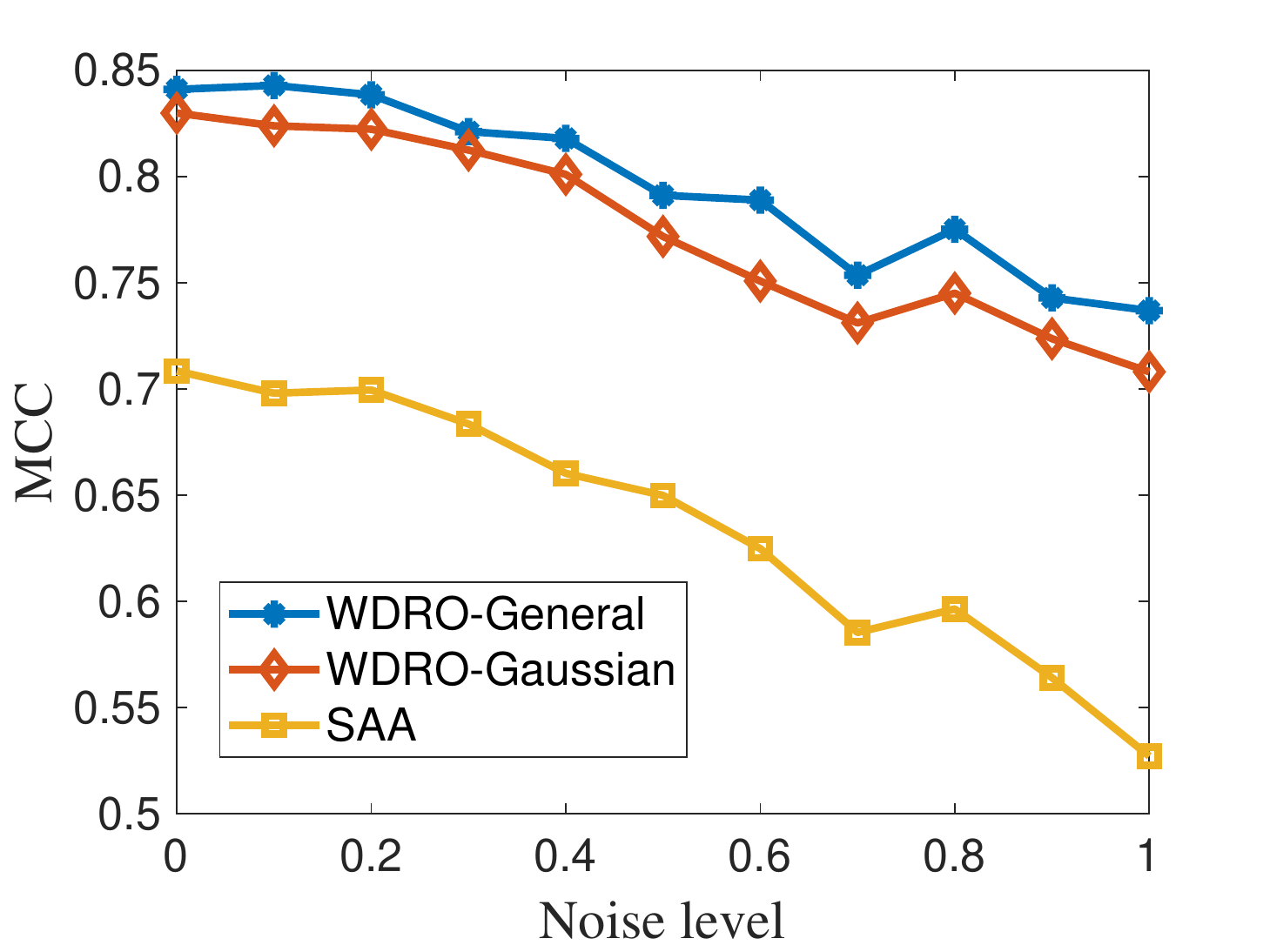}}
	  \quad
	  \subfloat[]{
        \includegraphics[width=0.48\linewidth]{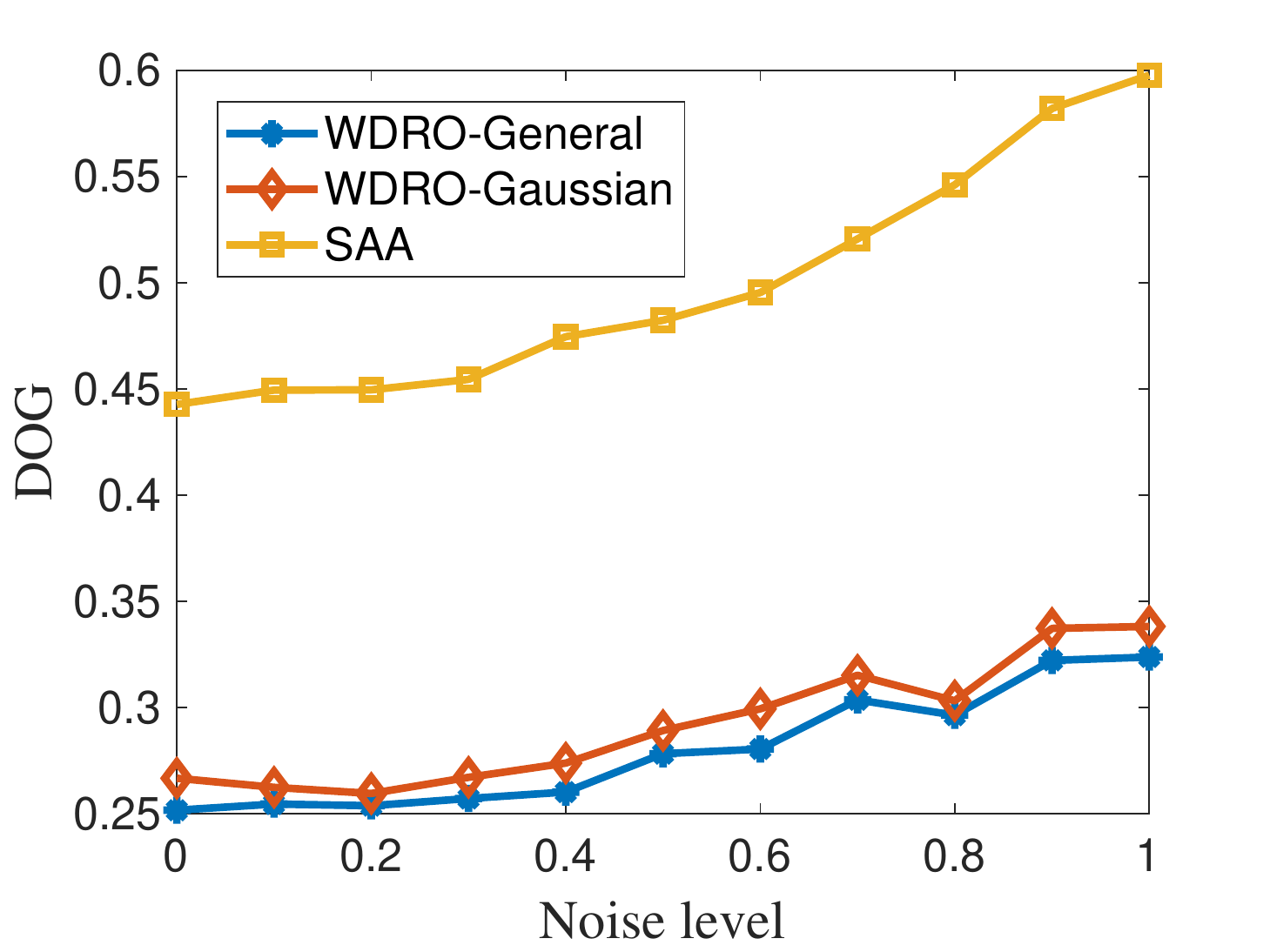}}
    	\caption{Performance of the learned graph with different levels of noise (a) MCC; (b) DOG }
    	\label{fig-noiselevel}
\end{figure}

In addition to the uncertainty caused by sample size, we also take noise-induced uncertainty into consideration. To this end, we fix the number of signals to 100 and change $\sigma_w$ from 0.1 to 1. As depicted in \ref{fig-noiselevel} (a), when noise level is low (lower than 0.5), it has little impact on MCC of the graph learned by WDRO framework while the performance of SAA drops significantly. When the noise level increases further, MCC of our framework also decreases but is still better than that of SAA. For another side, as shown in \ref{fig-noiselevel} (b), the impact of noise level on DOG is not as great as that of MCC. This may caused by that the misidentified edges, which can affect MCC considerably, have small weight. Hence, they have less impact on DOG than MCC.

\subsubsection{Impact of norm type $p$}
We then show the impact of norm type in cost function of Wasserstein distance. Since we assume $p=2$ in Gaussian scenario, we only take the general scenario into consideration. Additionally, for the reason that MCC and DOG show similar trend, we only list the results of MCC.
	\begin{table}[t]
		\centering
		\caption{MCC values of the learned graph using different norm type $p$ under general scenario}
		\begin{tabular}{cccccccc}
			\toprule
			Sam$p$  & 1  & 4/3  & 3/2 & 2 & 3 & 4 & $\infty$\\
			\midrule
			$N=50$   & 0.629 & 0.640   &0.656   & 0.704  & 0.736 & \textbf{0.740} & 0.627\\
    		$N=100$  & 0.706 & 0.723   &0.742   & 0.771  & 0.796 & \textbf{0.805} & 0.722\\
    		$N=200$  & 0.755 & 0.753   & 0.755  & 0.786  & 0.812 & \textbf{0.814} & 0.752\\
    		$N=1000$ & 0.820 & 0.822   & 0.824  & 0.832  &0.844  & \textbf{0.851} & 0.820\\
			\bottomrule
		\end{tabular}
		\label{table-MCC-p}
	\end{table}
	
As illustrated in Table \ref{table-MCC-p}, the performance of larger $p$ values is superior to those of small $p$ values except $p = \infty$. However, the superiority is not significant, and if take convenience into consideration, $p=2$ will be a suitable choice.

\subsubsection{Performance of pipeline tasks}
Next we check the performance of the learned graph on pipeline tasks. We apply Louvain algorithm \cite{blondel2008fast} to detect communities in the learned graph. For this purpose, we generate a SBM graph with 3 clusters and each cluster contains 15 vertices. The probabilities of node connections between clusters and within clusters are 0.02 and 0.3 respectively. To evaluate the performance of detection results, we adopt normalized mutual information (NMI) to measure the dependence between the cluster results of ground-truth and those of the learned graph. The results are shown in Table \ref{table-NMI}, from which we can see that in the case of small sample size, the performances of WDRO framework outperform those of SAA, illustrating that the detection results of our framework are more similar to the groundtruth. However, when the sample is large, the superiority of our framework is not as obvious as that of small size case because uncertainty decreases as sample size.

	\begin{table}[t]
		\centering
		\caption{Performance of community detection results using the learned graph}
		\begin{tabular}{ccccccc}
			\toprule
			$N$  & 80  & 100 & 150  & 200 & 500 & 1000 \\
			\midrule
			General   & 0.727          & 0.756           & \textbf{0.762}   & \textbf{0.814}  & \textbf{0.787} & 0.785 \\
    		Gaussian  & \textbf{0.737} & \textbf{0.7734}  &0.745   & \textbf{0.814} & 0.776 & \textbf{0.799} \\
    		SAA       & 0.700 & 0.699  & 0.725  & 0.774 & 0.779 & 0.781 \\
			\bottomrule
		\end{tabular}
		\label{table-NMI}
	\end{table}

\subsubsection{Efficiency of algorithms}
\begin{figure}[t] 
    \centering
       \includegraphics[width=0.48\linewidth]{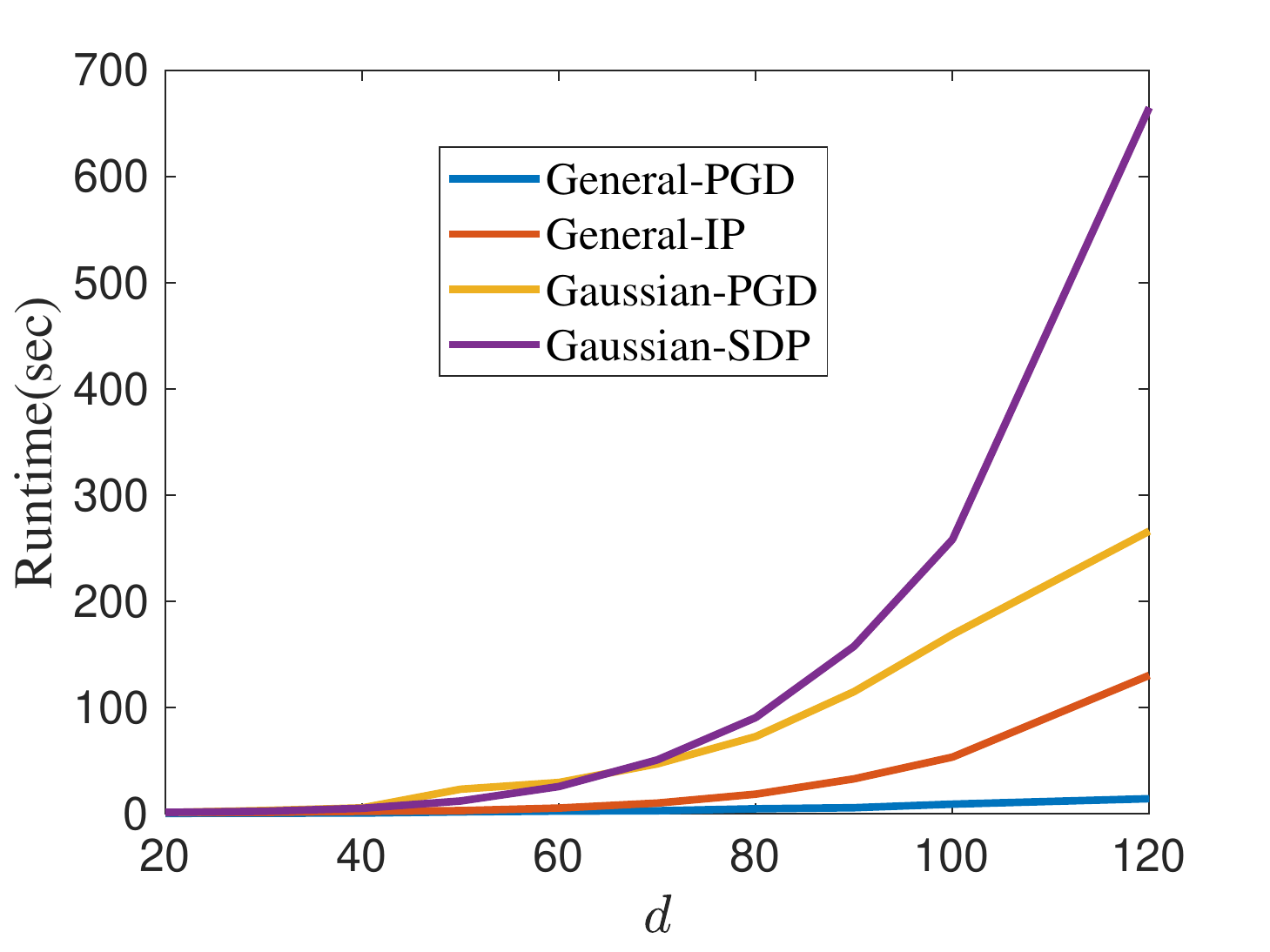}
    	\caption{Running time of different graph scales }
    	\label{fig-runtime-d}
\end{figure}

The last part of synthetic data experiments is testing the efficiency of the proposed algorithm. We compare our algorithm with classic interior point (IP) method . For Gaussian scenario, we use the use interior point to solve the SDP form of the origin problem. As depicted in \ref{fig-runtime-d}, the runtime of four algorithms all increases as the number of vertices in graph. However, as stated in complexity analysis part, the runtime of IP increases more drastically than that of our algorithm in both Gaussian and general scenarios. On the other hand, we can observe that Gaussian scenario is more time-consuming than general scenario since in Gaussian the algorithm costs order $\mathcal{O}(d^3)$ flops. Therefore, to reduce runtime, we can adopt some tricks to avoid these time-consuming operators. The details can be found in complexity analysis section. 

Another important fact is that the most time consuming scenario is SDP reformulation solved by interior point method.  This is due to the fact that SDP problem bring extra variables and constraints just as analyzed in section \ref{sec:formulation-sdp}. Actually, the reason we set the maximum of $d$ in Figure \ref{fig-runtime-d} 120 is that out-of-memory errors will occur if $d>150$ for the experiments of SDP problems. Therefore, it is not practical to reformulate ($\ref{basic infsup formulation}$) as a SDP problem.

\subsection{Real data}
We first apply our framework to the temperature data of 31 provincial capitals in Mainland China to learn a climate correlation graph between these cities. We collect temperature information from 2017 to 2019 \footnote{The data is available at website http://www.weather.com.cn/} and average the data of the first and second half month. Finally, 72 signals are obtain,the number of which is small for a 31-vertices graph, which brings large uncertainty in the collected data. Since no groundtruth graph is available, we only discuss the rationality of the learned graph.

\begin{figure}[t] 
    \centering
	  \subfloat[]{
       \includegraphics[width=0.48\linewidth]{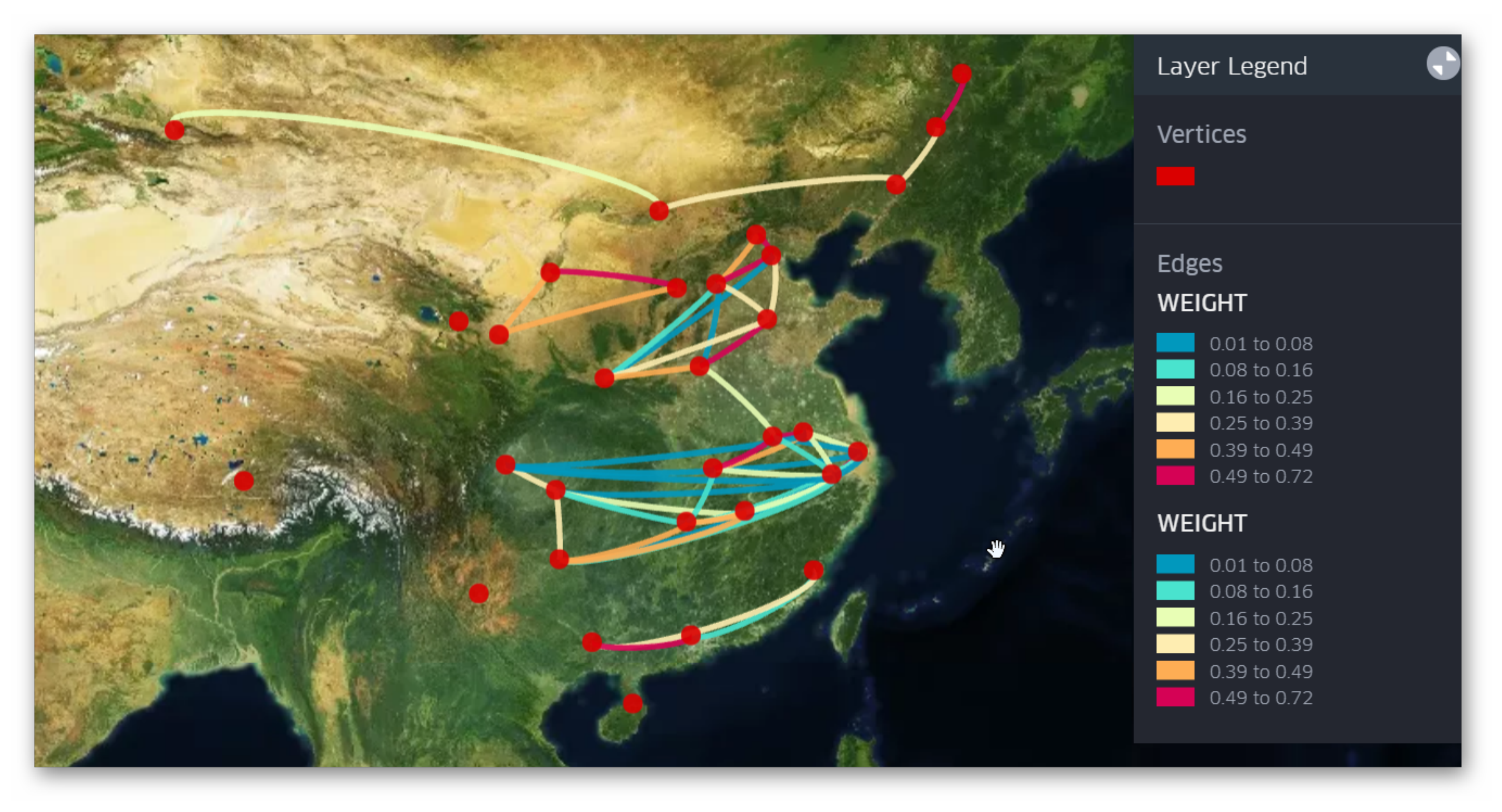}}
	  \quad
	  \subfloat[]{
        \includegraphics[width=0.48\linewidth]{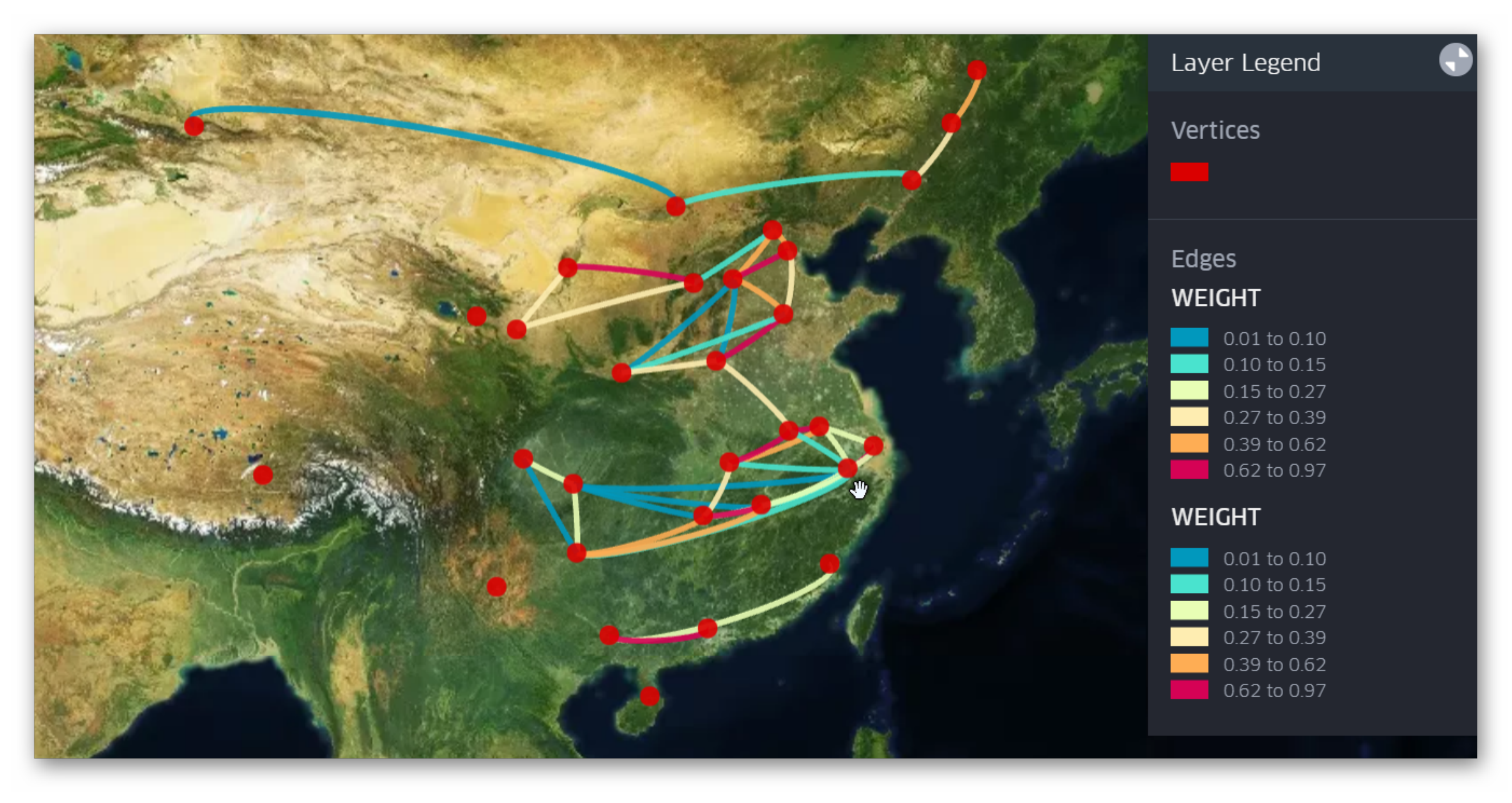}}
    	\caption{The learned temperature relationship graph of the mainland of China (a) general scenario; (b) Gaussian scenario }
    	\label{fig-temp}
\end{figure}

From Fig.\ref{fig-temp} (a) and (b), we can see that edges almost lie in latitude direction. It is reasonable because the temperatures in the same latitude area tend to be similar. If the latitude difference between two regions is large, then temperature difference between them will also be large, meaning a weaker temperature connection. Visually, these connections divide Chinese mainland into 4 regions from north to south. These 4 regions roughly correspond to cold-temperate/medium-temperate regions, warm-temperate regions, subtropical regions and tropical regions. It is also worth noting that there are four isolated points on the graph of general scenario, which are Xining, Lhasa, Kunming and Haikou. The first three cities are all in plateau, which means that their altitudes are much higher than other cities. High altitude makes them completely different from other regions climatically. The last isolated city is Haikou, which lies in a island in tropical region. It is reasonable to differ from other cities on land. 

\begin{figure}[t] 
    \centering
	  \subfloat[]{
       \includegraphics[width=0.48\linewidth]{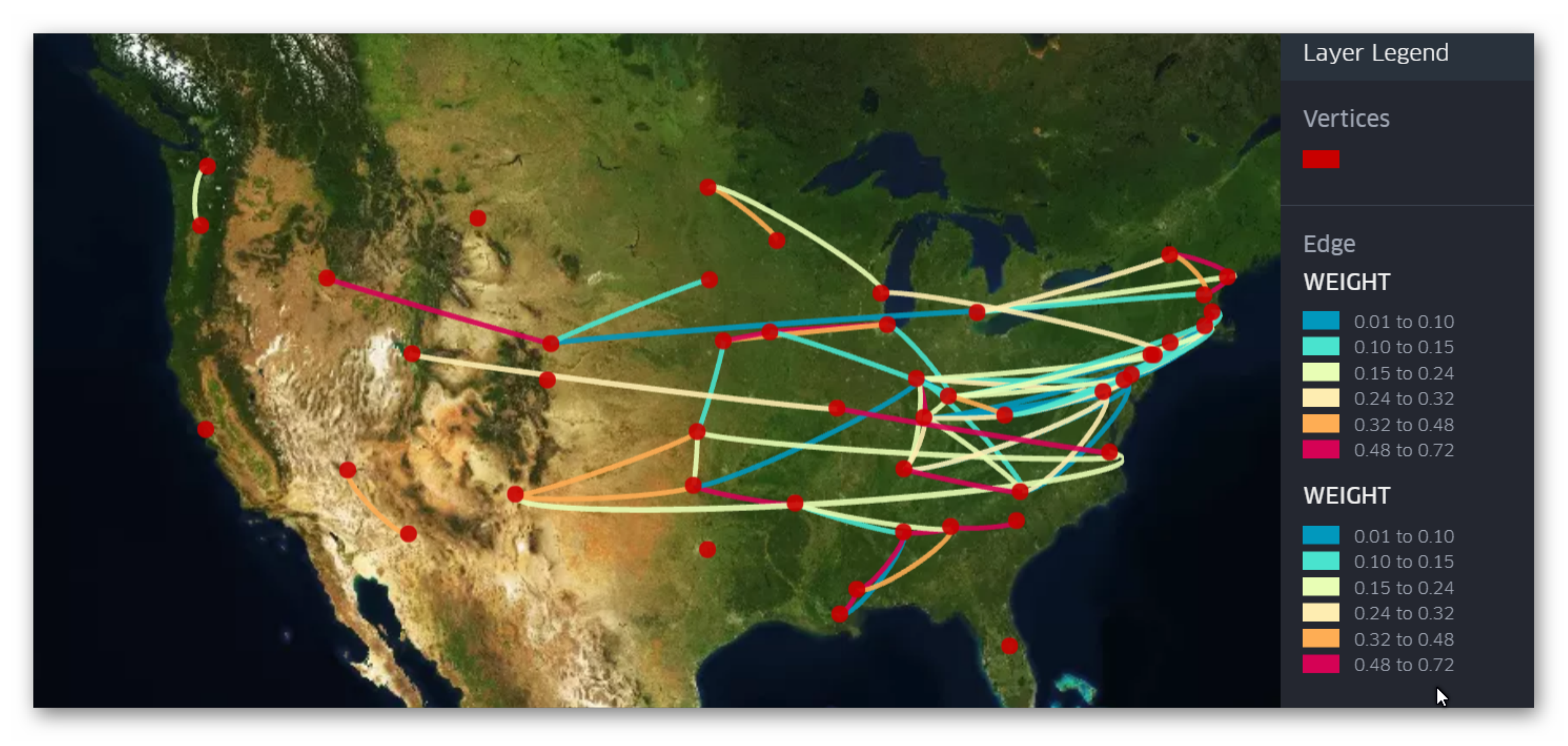}}
	  \quad
	  \subfloat[]{
        \includegraphics[width=0.48\linewidth]{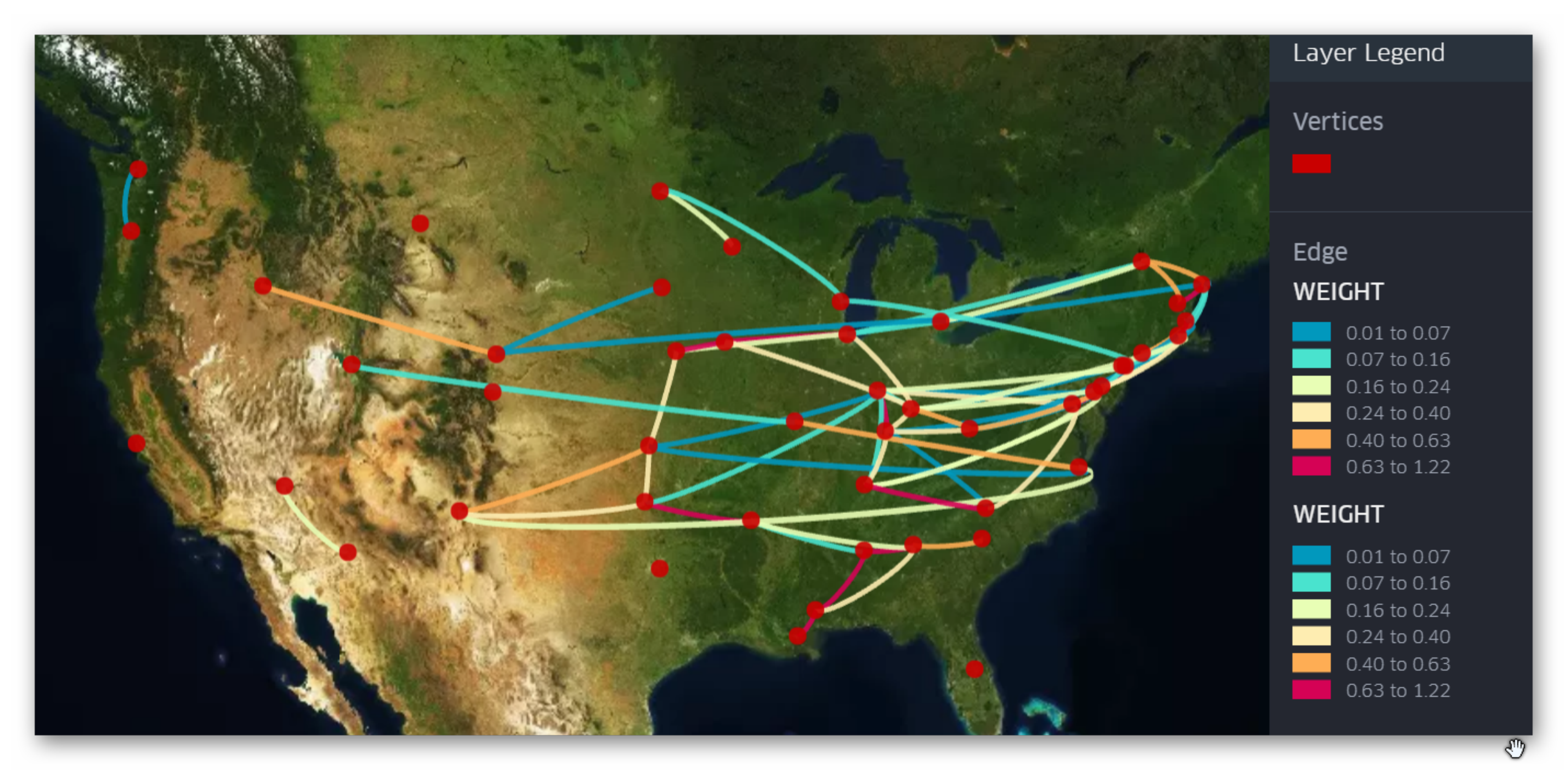}}
    	\caption{The learned temperature relationship graph of the mainland of the USA (a) general scenario; (b) Gaussian scenario }
    	\label{fig-temp-USA}
\end{figure}

We also apply our framework to temperature data of states of the mainland of USA except Alaska and Hawaii. Temperature data of 48 states are collected and for each states, we collected the weekly average temperature for 2020 and part of 2019 \footnote{The data is available at website https://data.iimedia.cn/}. Hence, a total of 60 signals for each state are collected. Note that 60 signals for a 48 vertices means high level of uncertainty. Same as the case above, we discuss the rationality of the learned graph. The results is shown in Fig.\ref{fig-temp-USA}. Same as the learned graph of China, edges in Fig.\ref{fig-temp-USA} almost lie in latitude direction. The 
isolated points are in high altitude or in low latitude, \textit{i.e.}, Florida. Another interesting trend is that edges in the east are much more those in the west. This may be caused by the fact plains are the most common terrain in  the east of the USA while the terrains in the west are
more complex. Furthermore, the distances between states in the west are also greater, which may cause greater climatic differences. The learned graphs in both general scenario and Gaussian scenario are reasonable, unveiling the power of our framework.

\section{Conclusion}
\label{sec:Conclusion}
In this paper, we are committed to the problem of learning a graph directly from data in the context of uncertainty. To this end, we propose a graph learning framework based on  Wasserstein distributionally robust optimization, which handles uncertainty from the perspective of data distribution. Specifically, we develop two graph learning models based on Gaussian distribution assumption and a more general circumstance without any prior distribution assumption. Two algorithms are also put forward to solve these models, which is proven to be more efficient than classic interior point method. Experimental results show that our framework can learn a reliable graph under uncertainty.

Future research directions include applying distributionally robust optimization framework to other prior assumptions except smoothness, as well as a more efficient algorithm accommodate large-scale graphs.

\appendices

\section{Proof of Theorem \ref{therorem-gaussian reformulation}}
\label{sec:appendix-1}
By the dual form of $R(\mathbf{L})$ in Lemma \ref{proposition-gaussian formulation dual form}, we notice the maximization $h(\gamma,\mathbf{L})$ can be decoupled with $\mathbf{\Sigma}$ and $\bm{\mu}$ separately. Let's define
\begin{align}
 l_1(\mathbf{\Sigma})  &\triangleq \text{Tr}[\mathbf{\Sigma}(\mathbf{L} - \gamma \mathbf{I})] +  2\gamma\text{Tr} (\sqrt{\mathbf{\Sigma}_n^{\frac{1}{2}} \mathbf{\Sigma} \mathbf{\Sigma}_n^{\frac{1}{2}}} ),\\   l_2(\bm{\mu}) &\triangleq \text{Tr}(\mathbf{L}\bm{\mu}\bm{\mu}^{\text{T}}) -\gamma\parallel \bm{\mu} -\bm{\mu}_n\parallel_2^2. 
\end{align}
Then $h(\gamma,\mathbf{L})$ can be written as
\begin{equation}
    h(\gamma,\mathbf{L}) = \sup_{\mathbf{\Sigma} \succeq \mathbf{0}} l_1(\mathbf{\Sigma}) +\sup_{\bm{\mu}}l_2(\mu).
\end{equation}
Therefore, we can calculate the maximum value of $l(\mathbf{\Sigma},\bm{\mu})$ with respect to $\mathbf{\Sigma}$ and $\bm{\mu}$ separately.


Firstly, we focus on $l_1(\mathbf{\Sigma})$, and define $\mathbf{\Sigma}_n^{1/2} \mathbf{\Sigma} \mathbf{\Sigma}_n^{1/2} \triangleq \mathbf{M}^2$, it can be found that
\begin{equation}
\sup_{\mathbf{\Sigma} \succeq \mathbf{0}}l_1(\mathbf{\Sigma}) =\sup_{\mathbf{M} \succeq \mathbf{0}} \text{Tr}\left(\mathbf{M}^2 \mathbf{\Sigma}_n^{-\frac{1}{2}}(\mathbf{L} - \gamma \mathbf{I})\mathbf{\Sigma}_n^{-\frac{1}{2}}\right) + 2\gamma\text{Tr}(\mathbf{M}).\label{eqn:M}
\end{equation}
Since the \emph{r.h.s} of \eqref{eqn:M} is a quadratic form of positive semi-definite matrix $\mathbf{M}$, the maximum exists only when $\gamma\mathbf{I} \succeq \mathbf{L}$ and the maximizer is
\begin{equation}
\mathbf{M}^* = -\gamma \mathbf{\Sigma}_n^{\frac{1}{2}}(\mathbf{L} - \gamma \mathbf{I})^{-1}\mathbf{\Sigma}_n^{\frac{1}{2}}.
\end{equation}
Bring $\mathbf{M}^*$ to \eqref{eqn:M}, we can get 
\begin{equation}
\sup_{\mathbf{\Sigma} \succeq \mathbf{0}}l_1(\mathbf{\Sigma}) = -\gamma^2 \text{Tr}\left(\mathbf{\Sigma}_n^{\frac{1}{2}}(\mathbf{L} - \gamma \mathbf{I})^{-1}\mathbf{\Sigma}_n^{\frac{1}{2}}\right).\label{eqn:l_1}
\end{equation}

Next, we turn our attention on $l_2(\bm{\mu})$. Calculate the derivative of $l_2(\bm{\mu})$ and we assume $\gamma\mathbf{I} \succ \mathbf{L}$, the maximizer $\bm{\mu}^*$ equals to $\gamma(\gamma\mathbf{I}-\mathbf{L})^{-1}\bm{\mu}_n$. Bring it back to $l_2(\bm{\mu})$, it is easy to obtain
\begin{equation}
   \sup_{\bm{\mu}}l_2(\mu) = - \gamma \bm{\mu}_n^{\text{T}}\bm{\mu}_n + \gamma^2(\gamma\mathbf{I}-\mathbf{L})^{-1} \bm{\mu}_n^{\text{T}}\bm{\mu}_n.\label{eqn:l_2}  
\end{equation}

Combining \eqref{eqn:l_1} and \eqref{eqn:l_2}, we reach the maximization $h(\gamma, \mathbf{L})$, that is, 
\begin{align}
h(\gamma, \mathbf{L}) &= -\gamma^2 \text{Tr}\left(\mathbf{\Sigma}_n^{\frac{1}{2}}(\mathbf{L} - \gamma \mathbf{I})^{-1}\mathbf{\Sigma}_n^{\frac{1}{2}}\right) - \gamma \bm{\mu}_n^{\text{T}}\bm{\mu}_n + \gamma^2(\gamma\mathbf{I}-\mathbf{L})^{-1} \bm{\mu}_n^{\text{T}}\bm{\mu}_n  \\
&= - \gamma \bm{\mu}_n^{\text{T}}\bm{\mu}_n +\gamma^2 \text{Tr}((\gamma\mathbf{I}-\mathbf{L})^{-1} \mathbf{\Sigma}_x).
\end{align}

Finally, bring $h(\gamma, \mathbf{L})$ back to  \eqref{gaussian formulation dual form} in Lemma \ref{proposition-gaussian formulation dual form},  the worst case risk $R(\mathbf{L})$ is equivalent to
\begin{equation}
    \begin{aligned}
    R(\mathbf{L}) =  & \underset{\gamma\mathbf{I} \succ \mathbf{L}}{\text{inf}}\,\, \gamma[\epsilon^2 - \text{Tr}(\mathbf{\Sigma}_n)]- \gamma \bm{\mu}_n^{\text{T}}\bm{\mu}_n +\gamma^2 \text{Tr}((\gamma\mathbf{I}-\mathbf{L})^{-1} \mathbf{\Sigma}_x )+\eta\parallel\mathbf{L} \parallel_{F}^2\\
     =&\underset{\gamma\mathbf{I} \succ \mathbf{L}}{\text{inf}}\,\, \gamma [\epsilon^2 - \text{Tr}(\mathbf{\Sigma}_x)]+\gamma^2\text{Tr}((\gamma\mathbf{I}-\mathbf{L})^{-1} \mathbf{\Sigma}_x)+\eta\parallel\mathbf{L} \parallel_{{F}}^2.
    \end{aligned}
    \label{appendix - gaussian formulation worst case}
\end{equation}
With (\ref{appendix - gaussian formulation worst case}), we can reach the conclusion of Theorem \ref{therorem-gaussian reformulation}.


\section{Proof of Theorem \ref{theorem-regularize}}
\label{sec:appendix-2}
The outline of this proof is similar with Theorem 2.1 in \cite{cisneros2020distributionally}. According to  corollary \ref{corollary dual form Q}, we can rewrite (\ref{basic infsup formulation}) as:
\begin{equation}
    \begin{aligned}
    &\underset{\mathbf{L}\in \mathcal{L}}{\text{inf}}\, \underset{\mathbb{P}\in \mathcal{P}}{\text{sup}}\, \mathbb{E}_\mathbb{P} [\mathbf{x}^{\text{T}}\mathbf{L}\mathbf{x}+\eta\parallel\mathbf{L} \parallel_{\text{F}}^2] \\= 
    &\underset{\mathbf{L}\in \mathcal{L},\gamma\geq  0}{\text{inf}} \,\, \left\{ \gamma \epsilon^{\alpha} +\eta\parallel\mathbf{L} \parallel_{\text{F}}^2 + \frac{1}{N} \sum_{i=1}^N  \underset{\mathbf{U} \in {\mathbb{S}^d}} {\text{sup}} \bigg\{\text{Tr}(\mathbf{L}\mathbf{U})-\gamma \parallel \textbf{vec}(\mathbf{U}) - \textbf{vec}(\mathbf{\Theta}_i)\parallel_p^{\alpha}\bigg\}  \right\}\\
    \triangleq &\underset{\mathbf{L}\in \mathcal{L},\gamma\geq  0}{\text{inf}} \,\, \left\{ \gamma \epsilon^{\alpha} +\eta\parallel\mathbf{L} \parallel_{\text{F}}^2 + \frac{1}{N} \sum_{i=1}^N \varphi_i(\mathbf{L}) \right\}
    \end{aligned}
    \label{appendix-rewrite-basic-formulation}
\end{equation}
We first focus our attention on the inner $\text{sup}$ of the dual form (\ref{appendix-rewrite-basic-formulation}) and calculate the close form of $\varphi_i(\mathbf{L})$. Define $\mathbf{\Delta} \triangleq \mathbf{U} - \mathbf{\Theta}_i$, and then
\begin{equation}
\begin{aligned}
\varphi_i(\mathbf{L}) 
 = &\underset{\mathbf{\Delta} \in {\mathbb{S}^d}} {\text{sup}} \left\{ \text{Tr}(\mathbf{L}(\mathbf{\Delta} + \mathbf{\Theta}_i)) - \gamma \parallel {\textbf{vec}}(\mathbf{\Delta})  \parallel_p^{\alpha} \right\}\\
 =  &\underset{\mathbf{\Delta} \in {\mathbb{S}^d}} {\text{sup}} \left\{ \text{Tr}(\mathbf{L}\mathbf{\Delta}) - \gamma \parallel {\textbf{vec}}(\mathbf{\Delta})\parallel_p^{\alpha}  + \text{Tr}(\mathbf{L}\mathbf{\Theta}_i) \right\}\\
\end{aligned}
\label{appendix-dual-innersup-1}
\end{equation}
Since $\text{Tr}(\mathbf{L}\mathbf{\Delta}) = \textbf{vec}(\mathbf{L}) \textbf{vec}(\mathbf{\Delta})$, apply Hölder inequality and we get $\text{Tr}(\mathbf{L}\mathbf{\Delta}) \leq \parallel \textbf{vec}(\mathbf{L}) \parallel_q \parallel \textbf{vec}(\mathbf{\Delta}) \parallel_p$, where the equality holds only when $\text{Tr}(\mathbf{L}\mathbf{\Delta}) > 0$ and there exists constant $\lambda$ so that $|\mathbf{\Delta}_{ij}|^p = \lambda |\mathbf{L}_{ij}|^q$ with $\frac{1}{p}+\frac{1}{q}=1$. For convenience, we define a set $\mathcal{S}$ containing all $\mathbf{\Delta}$ 
making the equation hold, that is, $\mathcal{S} = \{ \mathbf{\Delta}:\,\, \text{Tr}(\mathbf{L}\mathbf{\Delta})>0,|\mathbf{\Delta}_{ij}|^p = \lambda |\mathbf{L}_{ij}|^q,  \mathbf{\Delta} \in{\mathbb{S}^d}\}$. Based on this setting, (\ref{appendix-dual-innersup-1}) can written as:
\begin{equation}
\begin{aligned}
\varphi_i(\mathbf{L}) &=\underset{\mathbf{\Delta} \in {\mathcal{S}}}{\text{sup}}\{ \parallel \textbf{vec}(\mathbf{L}) \parallel_q \parallel \textbf{vec}(\mathbf{\Delta}) \parallel_p -\gamma \parallel {\textbf{vec}}(\mathbf{\Delta})  \parallel_p^{\alpha} +\text{Tr}(\mathbf{L}\mathbf{\Theta}_i) \}\\
 & \triangleq  \text{Tr}(\mathbf{L}\mathbf{\Theta}_i) +\underset{\mathbf{\Delta} \in  {\mathcal{S}}}{\text{sup}}  h(\mathbf{\Delta})
\end{aligned}
\label{appendix-dual-innersup-2}
\end{equation}
What we need to do is to calculate the maximum of $h(\mathbf{\Delta})$, and we will calculate in two different situations, that is, $\alpha=1$ and $\alpha \geq 1$.

When $\alpha = 1$, $h(\mathbf{\Delta})$ is actually a linear function with $\parallel \textbf{vec}(\mathbf{\Delta})\parallel_p$ and the maximum of $h(\mathbf{\Delta})$, which is equals to 0, exists only when $\gamma \geq \parallel\textbf{vec}(\mathbf{L})\parallel_q$. In this scenario, $\varphi_i(\mathbf{L})$ equals to $\text{Tr}(\mathbf{L}\mathbf{\Theta}_i)$. Replace it back to (\ref{appendix-rewrite-basic-formulation}), we can reach:
\begin{equation}
    \begin{aligned}
    &\underset{\mathbf{L}\in \mathcal{L}}{\text{inf}}\, \underset{\mathbb{P}\in \mathcal{P}}{\text{sup}}\, \mathbb{E}_\mathbb{P} [\mathbf{x}^{\text{T}}\mathbf{L}\mathbf{x}+\eta\parallel\mathbf{L} \parallel_{\text{F}}^2] \\ 
    =&\underset{\mathbf{L}\in \mathcal{L}}{\text{inf}}\, \underset{\gamma\geq  \parallel\textbf{vec}(\mathbf{L})\parallel_q}{\text{inf}}\,\, \left\{ \gamma \epsilon^{\alpha}+\eta\parallel\mathbf{L} \parallel_{\text{F}}^2 + \frac{1}{N} \sum_{i=1}^N  \text{Tr}(\mathbf{L}\mathbf{\Theta}_i)  \right\}\\
    =&\underset{\mathbf{L}\in \mathcal{L}}{\text{inf}} \,\,\epsilon\parallel \textbf{vec}(\mathbf{L}) \parallel_q + \frac{1}{N} \sum_{i=1}^N  \text{Tr}(\mathbf{L}\mathbf{\Theta}_i)+\eta\parallel\mathbf{L} \parallel_{\text{F}}^2 \\
    =& \underset{\mathbf{L}\in \mathcal{L}}{\text{inf}}\,\, \text{Tr}(\mathbf{L}\mathbf{\Theta}_n)+\epsilon\parallel \textbf{vec}(\mathbf{L}) \parallel_q+\eta\parallel\mathbf{L} \parallel_{\text{F}}^2
    \end{aligned}
    \label{appendix-rewrite-putback1}
\end{equation}
    
When $\alpha >1 $, we need to calculate the derivate of $h(\mathbf{\Delta})$ to get the maximum. After the first and second derivative test, we obtain that when $\parallel \textbf{vec}(\mathbf{\Delta}) \parallel_p\, = \left(\frac{\parallel \textbf{vec}(\mathbf{L} ) \parallel_q}{\gamma \alpha}\right)^{\frac{1}{\alpha -1}}$, $h(\mathbf{\Delta})$ reaches its maximum. Bring the maximizer to (\ref{appendix-dual-innersup-2}) to calculate the maximum and we get:
\begin{equation}
\begin{aligned}
\varphi_i(\mathbf{L}) 
= \parallel \textbf{vec}(\mathbf{L}) \parallel_q^{\frac{\alpha}{\alpha -1 }}\left(\frac{1}{(\gamma\alpha)^{\frac{1}{\alpha - 1}}}  - \frac{\gamma}{(\gamma\alpha)^{\frac{\alpha}{\alpha-1}}}\right) + \text{Tr}(\mathbf{L}\mathbf{\Theta}_i)
\end{aligned}
\label{appendix-dual-innersup-3}
\end{equation}
In the same way, put the maximum into (\ref{appendix-rewrite-basic-formulation}):
\begin{equation}
    \begin{aligned}
    &\underset{\mathbf{L}\in \mathcal{L}}{\text{inf}}\, \underset{\mathbb{P}\in \mathcal{P}}{\text{sup}}\, \mathbb{E}_\mathbb{P} [\mathbf{x}^{\text{T}}\mathbf{L}\mathbf{x}+\eta\parallel\mathbf{L} \parallel_{\text{F}}^2] \\ 
    =&\underset{\mathbf{L}\in \mathcal{L}}{\text{inf}} \underset{\gamma\geq  0}{\text{inf}}\,\, \bigg\{ \gamma \epsilon^{\alpha} +\eta\parallel\mathbf{L} \parallel_{\text{F}}^2
    + \parallel \textbf{vec}(\mathbf{L}) \parallel_q^{\frac{\alpha}{\alpha -1 }}\left(\frac{1}{(\gamma\alpha)^{\frac{1}{\alpha - 1}}}  - \frac{\gamma}{(\gamma\alpha)^{\frac{\alpha}{\alpha-1}}}\right) + \text{Tr}(\mathbf{L}\mathbf{\Theta}_n)  \bigg\}\\
    \end{aligned}
    \label{appendix-rewrite-putback2}
\end{equation}
Similarly, we need to calculate the inner minimum with $\gamma$ of (\ref{appendix-rewrite-putback2}). After the first and second derivative test, we can obtain the minimizer $\gamma^* = \frac{\parallel \textbf{vec}(\mathbf{L}) \parallel_q}{\alpha\epsilon^{\alpha-1}}$. Bring $\gamma^*$ to (\ref{appendix-rewrite-putback2}), we finally obtain:
\begin{equation}
    \begin{aligned}
    &\underset{\mathbf{L}\in \mathcal{L}}{\text{inf}}\, \underset{\mathbb{P}\in \mathcal{P}}{\text{sup}}\, \mathbb{E}_\mathbb{P} [\mathbf{x}^{\text{T}}\mathbf{L}\mathbf{x}+\eta\parallel\mathbf{L} \parallel_{\text{F}}^2] \\ 
    =&\underset{\mathbf{L}\in \mathcal{L}}{\text{inf}}  \text{Tr}(\mathbf{L}\mathbf{\Theta}_n)+\epsilon\parallel \textbf{vec}(\mathbf{L}) \parallel_q+\eta\parallel\mathbf{L} \parallel_{\text{F}}^2
    \end{aligned}
    \label{appendix-rewrite-putback3}
\end{equation}

In both $\alpha = 1$ and $\alpha>1$, we can reach the conclusion of Theorem \ref{theorem-regularize}, and finally we complete the proof.

\section{Definition of operator $\mathcal{T}$}
\label{sec:appendix-3}
Given a vector $\mathbf{v}\in \mathbb{R}^{d(d-1)/2}_+$, linear operator $\mathcal{T}$ is used to convert $\mathbf{v}$ to a Laplacian matrix, that is, $\mathbf{v}\mapsto \mathcal{T}\mathbf{v} \in \mathbb{R}^{d\times d}$, where $\mathcal{T}\mathbf{v}$ satisfies Laplacian constraints($[\mathcal{T}\mathbf{v}]_{ij} = [\mathcal{T}\mathbf{v}]_{ji}\leq 0$, for $i\leq j$ and $[\mathcal{T}\mathbf{v}]\cdot \mathbf{1} = 0$). Based on this, the linear operator can be defined as \cite{kumar2020unified}:
\begin{equation}
    \begin{aligned}
      {[}\mathcal{T}\mathbf{v}{]}_{ij} = 
        \begin{cases}
            -v_{i+b_j} & i > j, \\
            [\mathcal{T}\mathbf{v}]_{ji} & i< j,  \\
            -\sum_{i\neq j} [\mathcal{T}\mathbf{v}]_{ij} & i =j,
        \end{cases}
    \end{aligned}
    \label{linear-operator}
\end{equation}
where $b_j = -j +\frac{j-1}{2}(2d-j)$

The adjoint operator $\mathcal{T}^*$ of $\mathcal{T}$ can then be derived , that is, $\langle \mathcal{T}\mathbf{v}, \mathbf{V} \rangle = \langle \mathbf{v}, \mathcal{T}^*\mathbf{V} \rangle$. Specifically, for a matrix $\mathbf{V}$, the adjoint operator $\mathcal{T}^*: \mathbf{V} \mapsto \mathcal{T}^*\mathbf{v}$ is defined as
	\begin{equation}
        \begin{aligned}
        {[}\mathcal{T}^*\mathbf{V}{]}_k = \mathbf{V}_{ii}-\mathbf{V}_{ij}-\mathbf{V}_{ji}+\mathbf{V}_{jj},
        \end{aligned}
        \label{adjoint-operator}
    \end{equation}
where $i,j \in \mathbb{Z}_{+}$ and $k = i-j+\frac{j-1}{2}(2p-j)$ with $i>j$.



\ifCLASSOPTIONcaptionsoff
  \newpage
\fi


\bibliographystyle{ieeetr}
\bibliography{references}


\end{document}